\documentclass[preprint,11pt]{elsarticle}
\usepackage{epstopdf}
\usepackage{caption}
\usepackage{setspace}
\usepackage{amssymb}
\usepackage{multirow}
\usepackage{amsthm}
\usepackage{booktabs}
\usepackage{amsmath}
\usepackage{amssymb}
\usepackage{mathrsfs}
\usepackage{longtable}
\usepackage{lscape}
\usepackage{url}
\usepackage{tabularx}
\usepackage{epsfig}
\usepackage{colortbl}
\usepackage{subfigure}
\usepackage{tikz,mathpazo}
\usepackage{graphicx}
\usepackage{natbib}
\usepackage[noend]{algpseudocode}
\usepackage{algorithm}
\usepackage{algorithmicx}

\newtheorem{definition}{Definition}[section]
\usetikzlibrary{shapes.geometric, arrows}
\biboptions{numbers,sort&compress}

\begin{document}
\begin{frontmatter}
\title{Combining conflicting ordinal quantum evidences utilizing individual reliability}
\author[address1]{Yuanpeng He}
\author[address1]{Xingbin Wang\corref{label2}} 
\author[address1]{Yipeng Xiao\corref{label2}} 
\author[address1]{Fuyuan Xiao \corref{label1}}

\address[address1]{School of Computer and Information Science, Southwest University, Chongqing, 400715, China}
\cortext[label1]{Corresponding author: Fuyuan Xiao, School of
Computer and Information Science, Southwest University, Chongqing,
400715, China. Email address: xiaofuyuan@swu.edu.cn,
doctorxiaofy@hotmail.com.}

\cortext[label2]{The contributions of the authors are equal.}
\begin{abstract}
How to combine uncertain information from different sources has been a hot topic for years. However, with respect to ordinal quantum evidences contained in information, there is no any referable work which is able to provide a solution to this kind of problem. Besides, the method to dispel uncertainty of quantum information is still an open issue. Therefore, in this paper, a specially designed method is designed to provide an excellent method which improves the combination of ordinal quantum evidences reasonably and reduce the effects brought by uncertainty contained in quantum information simultaneously. Besides, some actual applications are provided to verify the correctness and validity of the proposed method.
\end{abstract}
\end{frontmatter}
\section{Introduction}
It is very important to measure the degree of uncertainty contained in information provided, which has been a hot topic in recent years. Lots of researchers have made prominent contributions to the field. Many theories are developed to reduce uncertainties of information provided which solve the problem from different levels. The representative works related to this field can be listed as $Z$ numbers \cite{li2020newuncertainty, Xiao2019a, zadeh2011note, tian2020zslf, Kang2019}, Dempster-Shafer evidence theory and the extensions of it \cite{book, Dempster1967Upper, Zadeh1986, Deng2019b}, $D$ numbers \cite{IJISTUDNumbers, xiao2019multiple, LiubyDFMEA, Deng2020ScienceChina} and complex mass function \cite{Xiao2019b, Xiao2020b, Xiao2020generalizedbelieffunction} which provide separate categories of solutions in handling uncertain information given by different sources.  However, how to correctly and effectively combine quantum evidences is an open problem. Some cutting-edge relevant concepts have been proposed to handle information given in the form of quantum \cite{Gao2019quantummodel, Dai2020, Huang2019}. However, a crucial factor of events is ignored which is the sequence of incidents to take place. Some papers have already introduce the concept of order when managing uncertain information from different angles \cite{pan2020Probability, Song2019POWA, Feng2019}. As a result, the order of propositions lying in a frame of discernment is taken into consideration as an prominent innovation of the proposed work in this paper. 

More than that, there exists a lack of effective rule of combination of quantum evidences and are no any other relative works. In general, with respect to traditional management of uncertain information, lots of works have been done to give different views on how to alleviate effects brought by uncertainties included in information given. Due to the effectiveness of the works, the relevant theories and method are applied into some actual situations, like pattern classification \cite{8944285, Liu2019}, target recognition \cite{LiuF2020TFS, Pan2020, Han2018} and decision making \cite{Han2019, Li2019b, Fei2019b, Song2019a, Liao2019}. The main categories of the improved method can be roughly divided into two parts, namely improved rule of combination and optimized data model. Some meaningful researches about the former kind are designed to manage conflicts among evidences \cite{ DBLP:journals/isci/Yager87a, Feng2016, jiang2019novel, Song2018} works which utilize diverse related theory to improve effects in handling conflicting evidences. As for the latter categories, one of the most effective method is to designed a base function for original data to eliminate extreme values \cite{Wang2019, DBLP:journals/apin/JingT21, https://doi.org/10.1002/int.22366} to help produce intuitive results of judgments. Nevertheless, in the field of quantum, the relative rules of combination is missing. In this paper, a completely new rule based on multi-layered system of judgment utilizing divergence measure and similarity calculation is proposed. The proposed method ensures the accuracy and rationality of the results combined and retain a complete figure on the description of actual situations expressed by given quantum frame of discernment.

The rest of the passage is organized as follows. The section of preliminaries introduces some related concepts to help construct a thorough system of judgment. Besides, the details of proposed method is clearly illustrated in the section of proposed method. Moreover, four applications are provided to verify the correctness and validity of the proposed method. In the last, conclusions are made to summarize the contributions of the whole paper.

\section{Preliminaries}

In this section, lots of related concepts are briefly introduced. Some meaningful works have been completed and applied into many actual applications, which reflects underlying and prominent efficiency of existing theory in handling problems in disposing uncertainty in information \cite{mao2020DEMATELevidence, Liufang2019beliefentropy, pan2020multi, Zhao2020complex}.

\subsection{Quantum model of mass function \cite{Gao2019quantummodel}}
\begin{definition}(Quantum mass function)\end{definition}
	On the basis of the definition of the quantum frame of discernment, the quantum mass function can be defined as:
	\begin{equation}
	P(|A\rangle)=\sigma e^{j\alpha},
	\end{equation}
	
	The quantum mass function is also called quantum basic probability assignment (QBPA) which is a mapping from P($\Theta$) to [0,1], which is defined as:
	\begin{equation}
	P(\phi)=0
	\end{equation}
	
	\begin{equation}
	\sum_{A\in P(\Theta)}|P(|A\rangle)|=1,
	\end{equation}
	
	Where $|P(|A\rangle)|$ is euqal to $\sigma^{2}$. The degree of belief to the proposition $|A\rangle$ is shown by $|\sigma|^2$. Besides, $\alpha$ represents phase angle of $|A\rangle$.
	
	\textbf{Remark 1: }The quantum mass function is regarded as the same as classic mass function when the phase angle equals $0^{\circ}$.	
	
	\textbf{Remark 2: }The quantum mass function does not satisfy the property of additivity, which is defined as:	
		\begin{equation}
		|P(|A\rangle)+P(|B\rangle)|\neq|P(|A\rangle)|+|P(|B\rangle)|
	\end{equation}
	
\begin{definition}(Quantum Combination Rule)
	
\end{definition}

Suppose that there are some QBPAs given as $Q_{1}$, $Q_{2}$ and so on. The process of combination of QBPAs is defined as:

\begin{equation}
\left\{
\begin{array}{lr}
P(\emptyset)=0\\\\
P(|A\rangle)=\dfrac{1}{1-K} \sum_{|B\rangle\cap|C\rangle \cap ... =|A\rangle}	\prod_{1 \leq i\leq j\leq 2} P_{i}(|B\rangle)\times P_{j}(|C\rangle) \times ... , \ \ \ |A\rangle\neq \emptyset
\end{array}
\right.
\end{equation}

Where K is defined as:

\begin{equation}
K=\sum_{|B\rangle\cap|C\rangle \cap ... =\emptyset}P_{1}(|B\rangle)\times P_{2}(|C\rangle) \times ...
\end{equation}

The step of normalization is defined as:

\begin{equation}
|P(|A\rangle)|=\dfrac{|P(|A\rangle)|}{|P(|A\rangle)|+|P(|B\rangle)|+...+|P(|A\rangle,|B\rangle)|+...}
\end{equation}

$K$ is called a special quantum probability, and $|K|$ shows the  degree of conflict among the quantum evidences.

\section{Proposed method}

In order to measure the degree of difference among evidences given and properly combine provided evidences in an ordinal environment, multiple-dimensional measure standard is established to offer a satisfying solution to this problem.

\begin{definition}(Ordinal Quantum Frame Of Discernment)
	
\end{definition}
The ordinal quantum frame of discernment  is a set whose elements are associated in a certain order, which is defined as:

\begin{equation}
\Theta_{ordinal}= \{ M_{1},M_{2},...,M_{n} \} 
\end{equation}

The sequence of propositions is  denoted by superscripts. The elements in an ordinal frame of discernment satisfy the following properties:
\begin{itemize}
	\item For any element with a superscript $i$, it is supposed to be ascertained before the one with superscript $i+n$ and $n \geq 1$.
	\item The definition of proposition in an ordinal frame of discernment is exactly the same as the ones defined in the traditional quantum frame of discernment except order of elements. 
	\item The level of uncertainty of the whole system can be further confirmed in the process of determining one more proposition. 
\end{itemize}

\begin{definition}
	(The degree of similarity of QBPAs)
\end{definition}
In an ordinal quantum frame of discernment, the values of propositions are given in the form of quantum. To get an underlying relationship among propositions contained in the quantum frame of discernment, as a result, how to figure out the method to describe the similarity between them must be taken into consideration. In the field of quantum frame of discernment, to better present features of separate values of propositions, vector is introduced into the  computation of disposing uncertain information. Meanwhile, in order to simplify the process of calculation, all operations which are carried out in four quadrants are mirrored to the first quadrant. The intersecting part of the area of two vectors is regarded as the initial similarity. $P_{i}(|A\rangle)$ and $P_{j}(|A\rangle)$ are utilized to represent two propositions in the field of quantum. Analogously, two pairs of $P^{real}$ and $P^{image}$ mean the real parts and imaginary parts of mass in the form of quantum. In addition, $P_{i\_least}^{real}(|A\rangle)$ and $P_{j\_least}^{imag}(|A\rangle)$ are on behalf of the least real part and the imaginary part between two propositions. The level of intermediate similarity of two evidences can be defined as:

  \begin{equation}
	Sim_{1}^{inter}(Q_{i},Q_{j})=\sum_{p\in\Theta}\frac {P_{i\_least}^{real}(|A\rangle)\times P_{j\_least}^{imag}(|A\rangle) \times 2}{P_{i}^{real}(|A\rangle)\times P_{i}^{imag}(|A\rangle)+P_{j}^{real}(|A\rangle)\times P_{j}^{imag}(|A\rangle)}
\end{equation}

Then, the process of the normalization of the intermediate similarity is defined as:

\begin{equation}
	Sim_{1}(Q_{i},Q_{j})=\frac{Sim_{1}^{inter}(Q_{i},Q_{j})}{\sum_{i=1}^{n}\sum_{j=1}^{n}Sim_{1}^{inter}(Q_{i},Q_{j})}
\end{equation}

\subsection{Proposed measurements on differences of quantum evidence}
\begin{definition}
	(The proposed end to end distance between QBPAs)
\end{definition}

Suppose that the number of QBPAs which are in the quantum frame of discernment $\Theta$ is $n$. To show the degree of deviation between QBPAs, the proposed end to end distance between two QBPAs is defined as:

\begin{equation}
d(Q_{i},Q_{j})=\sum_{|A\rangle \in 2^{\Theta}}|P_{1}(|A\rangle)-P_{2}(|A\rangle)|
\end{equation}

The step of normalization is defined as:

\begin{equation}
d_{XP}(Q_{i},Q_{j})=\dfrac{d(Q_{i},Q_{j})}{\sum_{1\leq i\leq j \leq n}d(Q_{i},Q_{j})}
\end{equation}

After the step of normalization, the first level of judgement of the measure of differentiation is completed. However, only one measurement of differences of evidences is not enough. Therefore, more tools in indicating the degree of differentiae are required.

\begin{definition}
		(The proposed relative distance between QBPAs)
\end{definition}
Suppose that there are $n$ QBPAs in the quantum frame of discernment $\Theta$. Moreover, when disposing mass given in the form of quantum, a generalized formula to measure distance of fuzzy sets may be helpful to indicate the level of discrepancy of quantum evidences due to similar form of them. Therefore, the method to obtain distance between two QBPAs utilizing fuzzy divergence can be defined as:

\begin{equation}
d(Q_{i},Q_{j})=\frac{1}{2}[\sum_{|A\rangle \in 2^{\Theta}}P_{i}(|A\rangle)\lg (\dfrac{2P_{i}(|A\rangle)}{P_{i}(|A\rangle)+P_{j}(|A\rangle)})+\sum_{|A\rangle \in 2^{\Theta}}P_{j}(|A\rangle)\lg (\dfrac{2P_{j}(|A\rangle)}{P_{j}(|A\rangle)+P_{i}(|A\rangle)}]
\end{equation}

The larger values the $d(Q_{i},Q_{j})$ are, the higher degree of difference the two quantum evidences have. In order to illustrate the degree of difference between different quantum evidences, the distances obtained are expected to be put under one unified standard. The normalized value, $d_{WB}$, which can be defined as:

\begin{equation}
d_{WB}(Q_{i},Q_{j})=\dfrac{d(Q_{i},Q_{j})}{\sum_{1\leq i\leq j \leq n}d(Q_{i},Q_{j})}
\end{equation}

\subsection{The process of modification of the ordinal quantum evidence system}

On account of that the frame of discernment is ordinal, there exists a decisive relationship between the uncertainty of evidence system and the number and sequences of propositions. In ordinal quantum system, the number of multiple propositions are expected to affect the level of uncertainty of the given frame of discernment, more alike proposition means that the evidence system is more uncertain. Besides, the sequence of the propositions is also crucial, the propositions which occur in the first place are considered to have a overwhelming effect on the propositions that occur after them. As a result, in the ordinal frame of discernment, the number and sequences of propositions must be taken into consideration. On the basis of the definition of the quantum frame of discernment, assume the number of propositions contained in the ordinal frame of discernment is $m$ and the sequence of a proposition is regarded as $n+1$. The detailed process of getting modified values is defined as:

(1) The computational formula whose weights of every proposition is $m-n$, which can be expressed specifically as:

\begin{equation}
Weights_{p_{i}}=m-n_{i}
\end{equation}

(2) Original mass of every proposition is denoted by $Mass_{q_{i}}$. Therefore, the process of obtaining improved intermediate values of each proposition is defined as:

\begin{equation}
Value_{p_{i}}=Mass_{p_{i}}\times Weights_{p_{i}}
\end{equation}

(3) The step of normalization of improved intermediate values of each proposition is defined as:

\begin{equation}
Value_{p_{i}}^{result}=\frac{Value_{p_{i}}}{\sum_{i=1}^{n} Value_{p_{i}} } 
\end{equation}

 \subsection{The procedure of obtaining specifically weighted results of combination}

According to the definition provided above,  $d_{YP}(Q_{i},Q_{j}) $ and $d_{WB}(Q_{i},Q_{j})$ can be obtained. In the same manner, $Sim_{1}(Q_{i},Q_{j}) $ can be obtained, too. In the ordinal frame of discernment,  $d_{YP}(Q_{i},Q_{j})$ and $d_{WB}(Q_{i},Q_{j})$ represent the degree of difference between two evidences. Besides, in order to manifest actual situations of the ordinal frame of discernment, the circumstantial process of getting the specifically weighted results of combining every quantum evidence can be expressed as:

(1) The distances are regarded as the degree of difference between two propositions, the following preliminary modified expression for intermediate similarity has better accuracy and congruence in the quantum frame of discernment, which can be defined as:

\begin{equation}
Sim_{2}^{inter}(Q_{i},Q_{j})=(1-d_{YP}(Q_{i},Q_{j}))\times(1-d_{WB}(Q_{i},Q_{j})
\end{equation}

(2) The steps of the  normalization of the intermediate similarity is defined as:

\begin{equation}
	Sim_{2}(Q_{i},Q_{j})=\frac{Sim_{2}^{inter}(Q_{i},Q_{j})}{\sum_{i=1}^{n}\sum_{j=1}^{n}Sim_{2}^{inter}(Q_{i},Q_{j})}
\end{equation}

(3) According to the definition mentioned above, from which $Sim_{1}(Q_{i},Q_{j})$ can be obtained, considering the initial expression for calculating degree of similarity between two quantum evidence and some further improvements of it. The sum of the two kinds of similarity is of great significance in modification for determining the weight of each evidence in an ordered system of judgement, which can be defined as:

\begin{equation}
SIM(Q_{i},Q_{j})=Sim_{1}(Q_{i},Q_{j})+Sim_{2}(Q_{i},Q_{j})
\end{equation}

(4)The calculation formula for intermediate weight of an evidence $i$ is defined as:

\begin{equation}
Wgt_{i}^{evidence}=\sum_{j=1}^{n} SIM(Q_{i},Q_{j}) 
\end{equation}

(5)The step of normalization of final improved weight of an evidence $i$ is defined as:
\begin{equation}
Wgt_{i}^{nor}=\frac{Wgt_{i}^{evidence}}{\sum_{i=1}^{n}\sum_{j=1}^{n}SIM(Q_{i},Q_{j}) }
\end{equation}

(6) The process of obtaining final modified value for specific proposition $p$ is defined as:
\begin{equation}
	Value_{p}^{final}=\sum_{i=1}^{n} Wgt_{i}^{nor} \times Value_{p}^{result}
\end{equation}

(7) A step of normalization is designed to ensure the sum of the values of propositions is exactly equal to 1, which is defined as:
\begin{equation}
	Value_{p_{j}}^{final_{nor}} = \frac{Value_{p_{j}}^{final}}{\sum_{i = 1}^{n} Value_{p_{i}}^{final}}
\end{equation}

Finally, the eventual values of propositions are obtained to serve as a standard of judgments.
		
\section{Applications}

\subsection{Application of medical diagnosis} 

The dispose of quantum evidence is of great significance to medical diagnosis. How to make correct judgments to medical information is still an urgent issue. The method proposed in this paper is more effective than traditional methods in combination of evidence conflicts. The example in the following shows the advantages of method in medical diagnosis.

\begin{table}[h]\footnotesize
	\centering
	\caption{Quantum Evidences given by medical equipment}
	\begin{spacing}{1.80}
		\begin{tabular}{c c c c c }\hline
			$Evidences$ & \multicolumn{4}{c}{$Values \ \ of \ \ propositions$}\\\hline
			& $\{C\}$ & $\{F\}$ & $\{S\}$ & $\{CS\}$\\
			$Evidence_{1}$ & $0.7416e^{0.4882j}$ & $0.4472
			e^{0.3165j}$ & $0.3873
			e^{0.3410j}$ & $0.3162e^{0.1988j}$\\
			& $\{F\}$ & $\{C\}$ & $\{CS\}$ & $\{S\}$\\
			$Evidence_{2}$ & $0.6708e^{0.6476j}$ & $0.5000e^{0.3176j}$ & $0.4123e^{0.6307j}$ & $0.3607e^{0.6077j}$\\
			& $\{F\}$ & $\{C\}$ & $\{S\}$ & $\{CS\}$\\
			$Evidence_{3}$ & $0.7280e^{0.5774j}$ & $0.3873e^{0.3561j}$ & $0.3162^{0.5099j}$ & $0.4690e^{0.6408j}$\\
			& $\{CS\}$ & $\{S\}$ & $\{C\}$ & $\{F\}$\\
			$Evidence_{4}$ & $0.8062e^{0.4527j}$ & $0.3606e^{0.4007j}$ & $0.2828e^{0.4942j}$ & $0.3742e^{0.4735j}$\\
			\hline
		\end{tabular}
	\end{spacing}
	%\label{tab:Margin_settings}
	\label{w1}
\end{table}

Assume that there is a hospital which equips advanced diagnostic instruments. The machine can test physical characteristics through some sensors and analyse the information gathered automatically.

\begin{table}[h]\footnotesize
	\centering
	\caption{The calculation result of parameter $d_{XP}(Q_{i},Q_{j})$ of application 1}
	\begin{spacing}{1.80}
		\begin{tabular}{c c c c c c }\hline
			$Evidences$ & \multicolumn{4}{c}{$Values \ \ of \ \ distance \ \ between \ \ QBPAs $}\\\hline
			& $Evidence_{1}$ & $Evidence_{2}$ & $Evidence_{3}$ & $Evidence_{4}$ \\
			$Evidence_{1}$ & $ 0 $ & $ 0.180431287 $ & $ 0.179981299 $ & $ 0.150345454$ \\
			$Evidence_{2}$ & $ 0.180431287 $ & $0  $ & $ 0.159564419 $ & $ 0.209037016$ \\
			$Evidence_{3}$ & $ 0.179981299 $ & $ 0.159564419 $ & $ 0 $ & $ 0.120640525$ \\
			$Evidence_{4}$ & $ 0.150345454 $ & $ 0.209037016 $ & $ 0.120640525 $ & $ 0$ \\ \hline
			
		\end{tabular}
	\end{spacing}
	\label{w1XP}
	
\end{table}

Based on the situations discussed above, a basic quantum frame of discernment of the specific problems is defined as $\Theta=\{C,F,S,CS\}$. Cough is denoted by $C$ and  fever is presented by $F$. In addition, stomachache is indicated by $S$ and cough and stomachache is denoted by $CS$. Details about the quantum evidence is given in Table \ref{w1}.

\begin{figure}[h] %figure环境，h默认参数是可以浮动，不是固定在当前位置。如果要不浮动，你就可以使用大写float宏包的H参数，固定图片在当前位置，禁止浮动。
	\centering %使图片居中显示
	\includegraphics[width=0.7\textwidth]{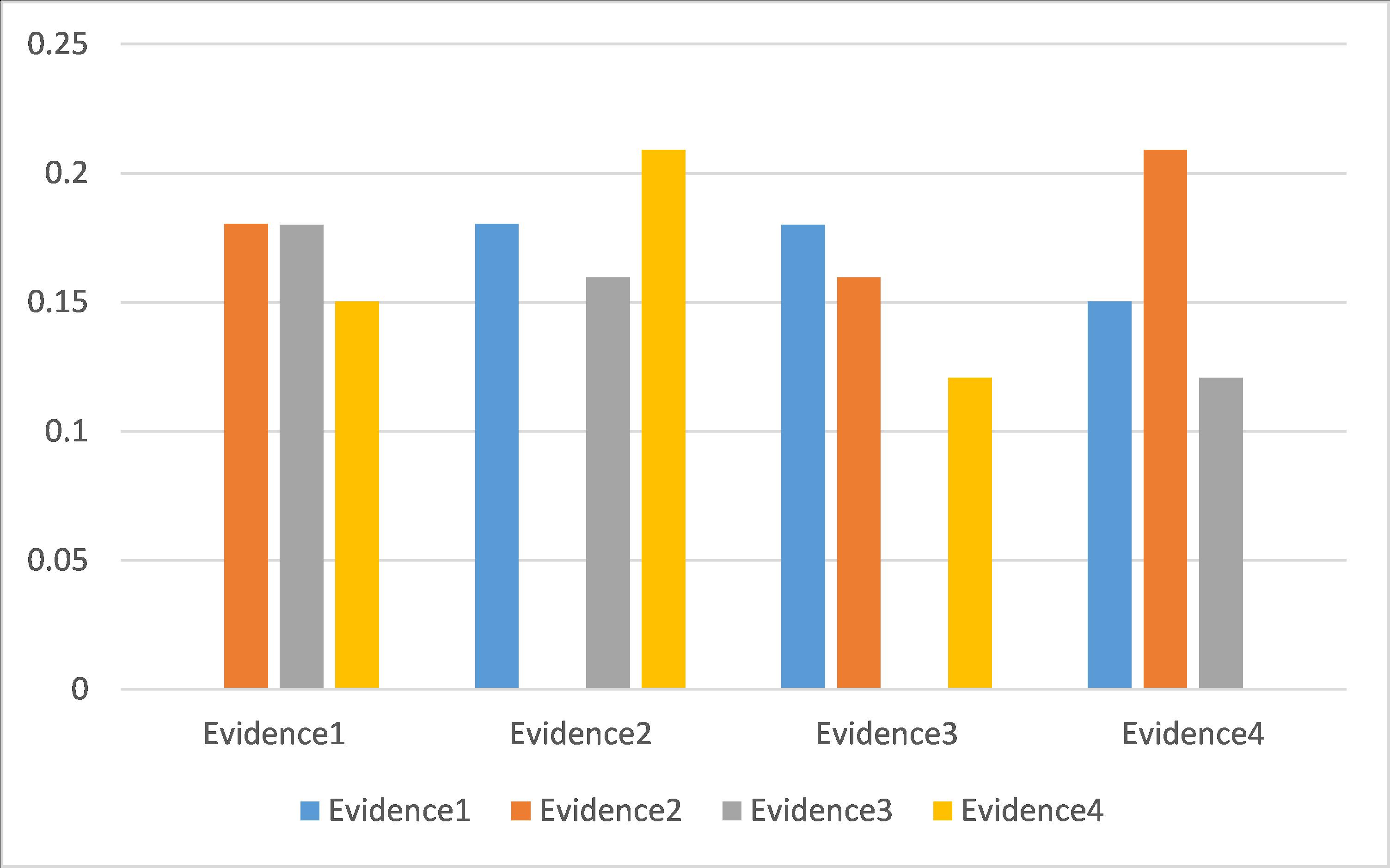} %中括号中的参数是设置图片充满文档的大小，你也可以使用小数来缩小图片的尺寸。
	\caption{The calculation result of parameter $d_{XP}(Q_{i},Q_{j})$ of application 1} %caption是用来给图片加上图题的
	\label{1} %这是添加标签，方便在文章中引用图片。
\end{figure}

Then, to output more valuable evidences to ensure what kind of symptom did the patient have, some other work is done. Firstly, the results of calculation of $d_{XP}(Q_{i},Q_{j})$ and $d_{WB}(Q_{i},Q_{j})$ are shown in Table \ref{w1XP} and Table \ref{w1WB}. Then, $Sim_{1}(Q_{i},Q_{j})$ and $ Sim_{2}(Q_{i},Q_{j})$ are obtained by utilizing regulation of the similarity and two kinds of
\begin{table}[h]\footnotesize
	\centering
	\caption{The calculation result of parameter $d_{WB}(Q_{i},Q_{j})$ of application 1}
	\begin{spacing}{1.80}
		\begin{tabular}{c c c c c c }\hline
			$Evidences$ & \multicolumn{4}{c}{$Values \ \ of \ \ distance \ \ between \ \ QBPAs $}\\\hline
			& $Evidence_{1}$ & $Evidence_{2}$ & $Evidence_{3}$ & $Evidence_{4}$ \\
			$Evidence_{1}$ & $ 0 $ & $ 0.138404579 $ & $ 0.163484138 $ & $ 0.210720747$ \\
			$Evidence_{2}$ & $ 0.138404579 $ & $0  $ & $ 0.117352038 $ & $ 0.170080205$ \\
			$Evidence_{3}$ & $ 0.163484138 $ & $ 0.117352038 $ & $ 0 $ & $ 0.199958292$ \\
			$Evidence_{4}$ & $ 0.210720747 $ & $ 0.170080205 $ & $ 0.199958292 $ & $0$  \\ \hline
		\end{tabular}
	\end{spacing}
	\label{w1WB}
\end{table}
distances of quantum evidences and are listed in Table \ref{w1Sim1} and Table \ref{w1Sim2}. Besides, the weight of each evidence provided by diagnostic instruments is acquired by combining two types of similarities mentioned before
\begin{figure}[h] %figure环境，h默认参数是可以浮动，不是固定在当前位置。如果要不浮动，你就可以使用大写float宏包的H参数，固定图片在当前位置，禁止浮动。
	\centering %使图片居中显示
	\includegraphics[width=0.7\textwidth]{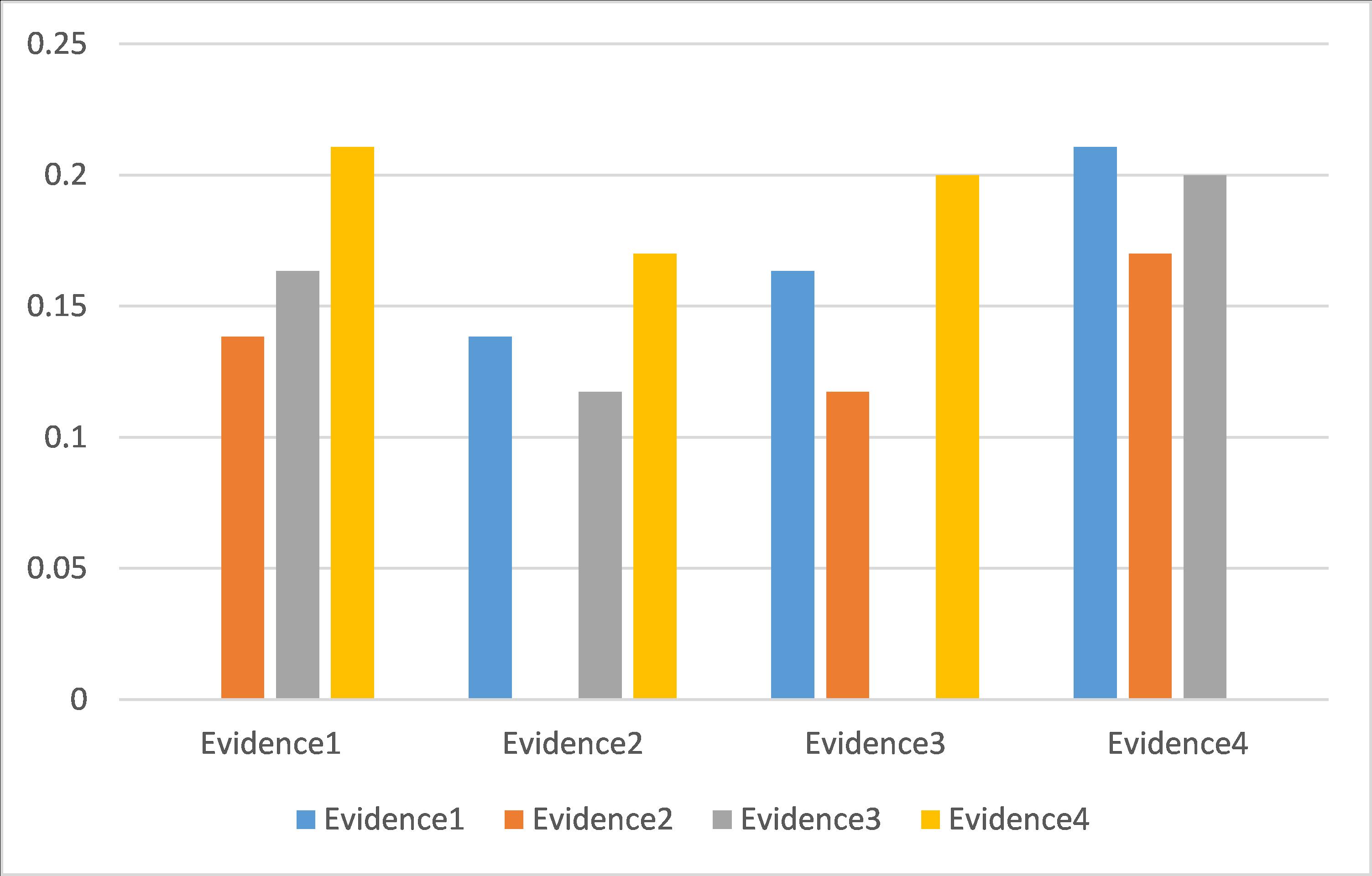} %中括号中的参数是设置图片充满文档的大小，你也可以使用小数来缩小图片的尺寸。
	\caption{The calculation result of parameter $d_{WB}(Q_{i},Q_{j})$ of application 1} %caption是用来给图片加上图题的
	\label{1} %这是添加标签，方便在文章中引用图片。
\end{figure}
 which are presented in Table \ref{w1weights}. In the last, the comparison of results which are combined by making use of proposed method and traditional rule of combination on the level at quantum and classic probability assignment are presented in Table \ref{w1quantumvalue} and Table \ref{w1value} respectively.

\begin{table}[h]\footnotesize
	\centering
	\caption{The calculation result of parameter $Sim_{1}(Q_{i},Q_{j})$ of application 1}
	\begin{spacing}{1.80}
		\begin{tabular}{c c c c c }\hline
			$Quantum \ \ evidence $ &\multicolumn{4}{c}{ $Values \ \ of \ \ similarity \ \ between \ \ QBPAs $}\\\hline
			
			& $Evidence_{1}$ & $Evidence_{2}$ & $Evidence_{3}$ & $Evidence_{4}$  \\
			$Evidence_{1}$ & $  1 $ & $ 0.221107846 $ & $ 0.169240021 $ & $ 0.149963075$ \\
			$Evidence_{2}$ & $ 0.221107846 $ & $ 1 $ & $ 0.176609694 $ & $ 0.142552885 $\\
			$Evidence_{3}$ & $ 0.169240021 $ & $ 0.176609694 $ & $ 1 $ & $ 0.140526479$ \\
			$Evidence_{4}$ & $ 0.149963075 $ & $ 0.142552885 $ & $ 0.140526479 $ & $ 1$ \\ \hline
		\end{tabular}
	\end{spacing}
	\label{w1Sim1}
	
\end{table}
It’s easily concluded from both of raw evidences and modified evidences that the patient gets a cough. By contrastive analysis, the evidence from four sensors is quite similar which leads that the weight of each evidence provided by medical diagnosis instruments is almost same. Also, Table\ref{w1value} shows that results of the combination of raw evidences and modified evidences are
\begin{figure}[h] %figure环境，h默认参数是可以浮动，不是固定在当前位置。如果要不浮动，你就可以使用大写float宏包的H参数，固定图片在当前位置，禁止浮动。
	\centering %使图片居中显示
	\includegraphics[width=0.7\textwidth]{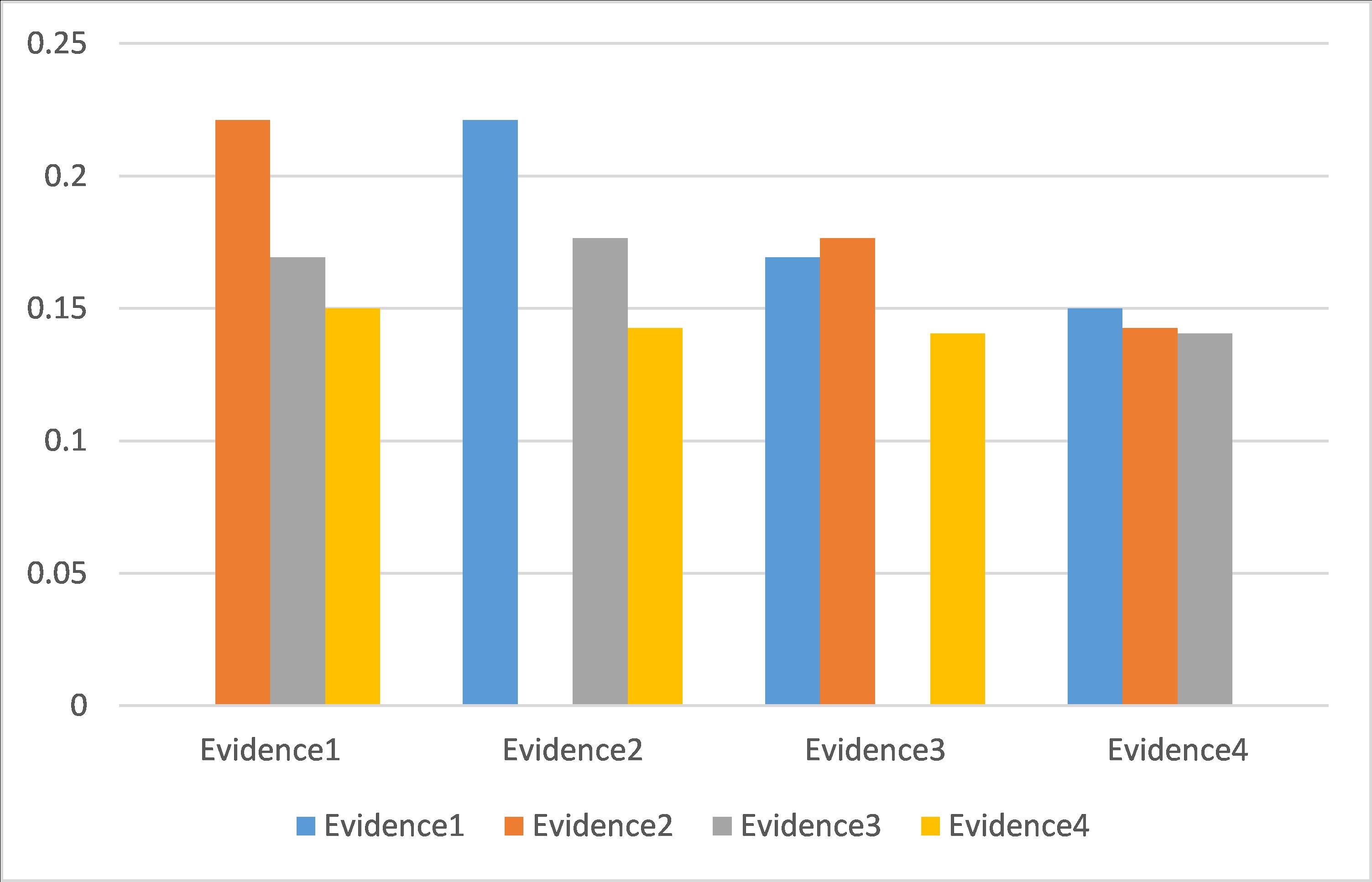} %中括号中的参数是设置图片充满文档的大小，你也可以使用小数来缩小图片的尺寸。
	\caption{The calculation result of parameter $Sim_{1}(Q_{i},Q_{j})$ of application1} %caption是用来给图片加上图题的
	\label{1} %这是添加标签，方便在文章中引用图片。
\end{figure}
nearly the same in the medical diagnosis. It indicates that when all evidences believe that a proposition is the most possible to happen, the combination taking the order of proposition into account can output results consistent with the ones produced by the traditional process of combination. The cause of the result is that although the order of propositions is taken into account, almost all possible ordering of the four propositions emerges. Thus, the order doesn't work in the process of combination.

\begin{table}[h]\footnotesize
	\centering
	\caption{The calculation result of parameter $Sim_{2}(Q_{i},Q_{j})$ of application 1}
	\begin{spacing}{1.80}
		\begin{tabular}{c c c c c }\hline
			$Quantum \ \ evidence $ & \multicolumn{4}{c}{ $Values \ \ of \ \ similarity \ \ between \ \ QBPAs $ }\\\hline
			
			& $Evidence_{1}$ & $Evidence_{2}$ & $Evidence_{3}$ & $Evidence_{4}$  \\
			$Evidence_{1}$ & $ 1 $ & $ 0.181458539 $ & $ 0.176105299 $ & $ 0.191831419$ \\
			$Evidence_{2}$ & $ 0.181458539 $ & $ 1 $ & $ 0.190480753 $ & $ 0.171415843$ \\
			$Evidence_{3}$ & $ 0.176105299 $ & $ 0.190480753 $ & $ 1 $ & $ 0.088708148$ \\
			$Evidence_{4}$ & $ 0.191831419 $ & $ 0.171415843 $ & $ 0.088708148 $ & $ 1$ \\ \hline
		\end{tabular}
	\end{spacing}
	%\label{tab:Margin_settings}
	\label{w1Sim2}
\end{table}
\begin{figure}[h] %figure环境，h默认参数是可以浮动，不是固定在当前位置。如果要不浮动，你就可以使用大写float宏包的H参数，固定图片在当前位置，禁止浮动。
	\centering %使图片居中显示
	\includegraphics[width=0.7\textwidth]{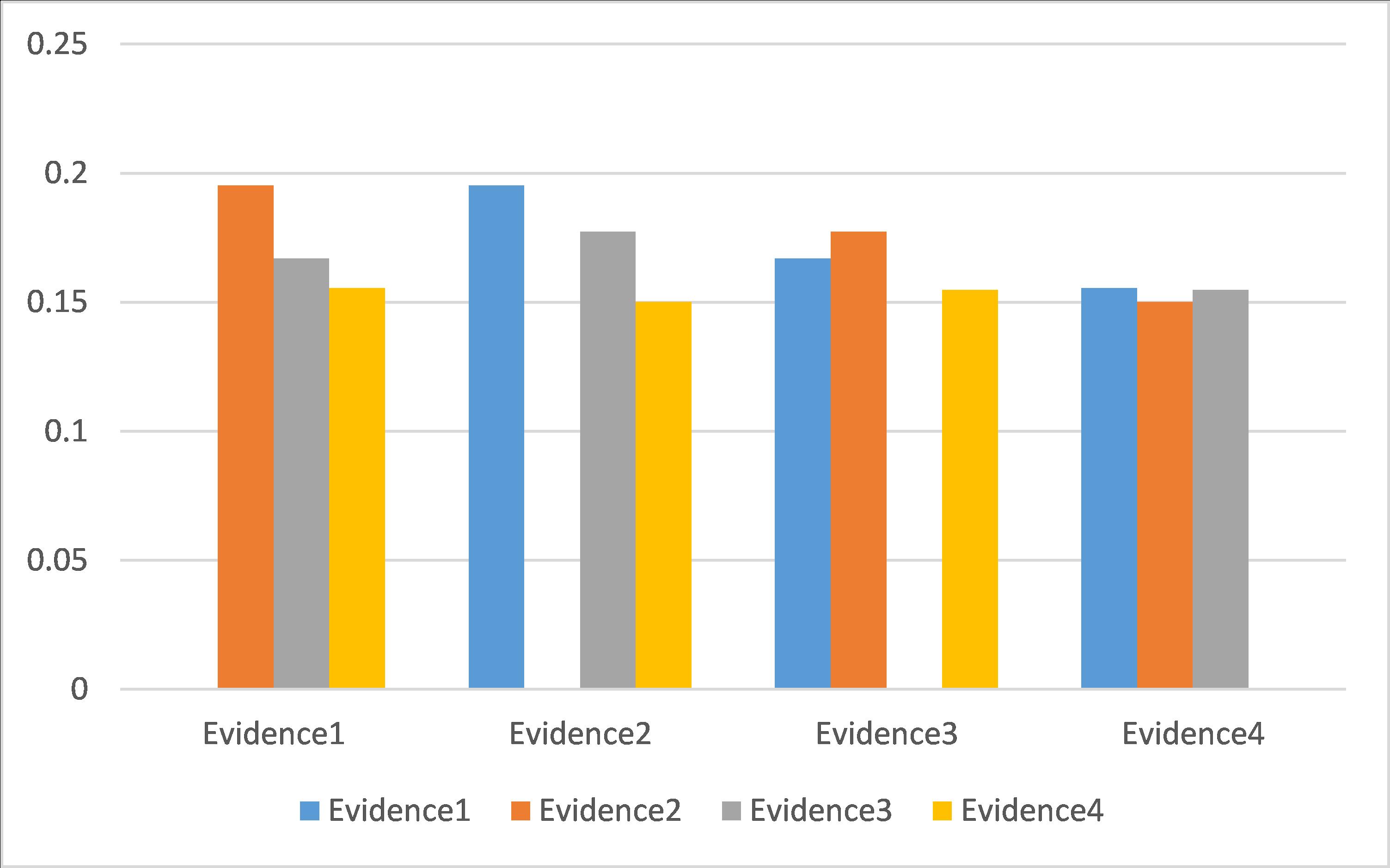} %中括号中的参数是设置图片充满文档的大小，你也可以使用小数来缩小图片的尺寸。
	\caption{The calculation result of parameter $Sim_{2}(Q_{i},Q_{j})$ of application 1} %caption是用来给图片加上图题的
	\label{1} %这是添加标签，方便在文章中引用图片。
\end{figure}
\begin{table}[h]\footnotesize
	\centering
	\caption{The calculation result of parameter $Wgt_{i}^{nor}$ of application1}
	\begin{spacing}{1.80}
		\begin{tabular}{c c c c c }\hline
			
			$Quantum \ \ evidence $ & $Evidence_{1}$ & $Evidence_{2}$ & $Evidence_{3}$ & $Evidence_{4}$  \\ \hline
			$WEIGHT(NOR) $ & $ 0.2724 $ & $ 0.2709 $ & $ 0.2354 $ & $ 0.2212$  \\ \hline
		\end{tabular}
	\end{spacing}
	%\label{tab:Margin_settings}
	\label{w1weights}
\end{table}
\begin{table}[h]\footnotesize
	\centering
	\caption{The comparison of results combined between proposed method and traditional rule of combination in quantum field in application 1 }
	\begin{spacing}{1.80}
		\begin{tabular}{c c c c c c}\hline
			
			$	Proposition $ & $ S $ & $ F $ & $ S $ & $ CS$ \\ \hline
			$	Values \ \ of \ \ modified \ \ combination $ & $0.9927e^{-2.4640i}$ & $0.1170e^{3.0032i}$ & $0.0189e^{-2.6956i}$ & $0.0229e^{-2.6187i}$ \\ \hline
			$Proposition $ & $ S $ & $ F $ & $ S $ & $ CS$ \\ \hline
			$	Values \ \ of\ \ basic\ \ combination $ & $0.9910e^{-2.41972i}$ & $0.1047e^{3.0943i}$ & $0.0479e^{-2.6047i}$ & $0.0671e^{-2.6600i}$ \\ \hline
		\end{tabular}
	\end{spacing}
	%\label{tab:Margin_settings}
	\label{w1quantumvalue}
	
\end{table}
\begin{table}[h]\footnotesize
	\centering
	\caption{The comparison of results combined between proposed method and traditional rule of combination in quantum field in application 1}
	\begin{spacing}{1.80}
		\begin{tabular}{c c c c c c}\hline
			
			$	Proposition $ & $ S $ & $ F $ & $ S $ & $ CS$  \\ \hline
			$	Values\ \ of\ \ modified\ \ combination $ & $ 0.9854$ & $ 0.0137 $ & $ 0.0003 $ & $ 0.0005$    \\ \hline
			$	Proposition $ & $ S $ & $ F $ & $ S $ & $ CS$ \\ \hline
			$	Values\ \ of\ \ basic\ \ combination $ & $ 0.9822$ & $ 0.0109 $ & $ 0.0023 $ & $ 0.0045  $  \\ \hline
		\end{tabular}
	\end{spacing}
	%\label{tab:Margin_settings}
	\label{w1value}
\end{table}						
\begin{figure}[h] %figure环境，h默认参数是可以浮动，不是固定在当前位置。如果要不浮动，你就可以使用大写float宏包的H参数，固定图片在当前位置，禁止浮动。
	\centering %使图片居中显示
	\includegraphics[width=0.7\textwidth]{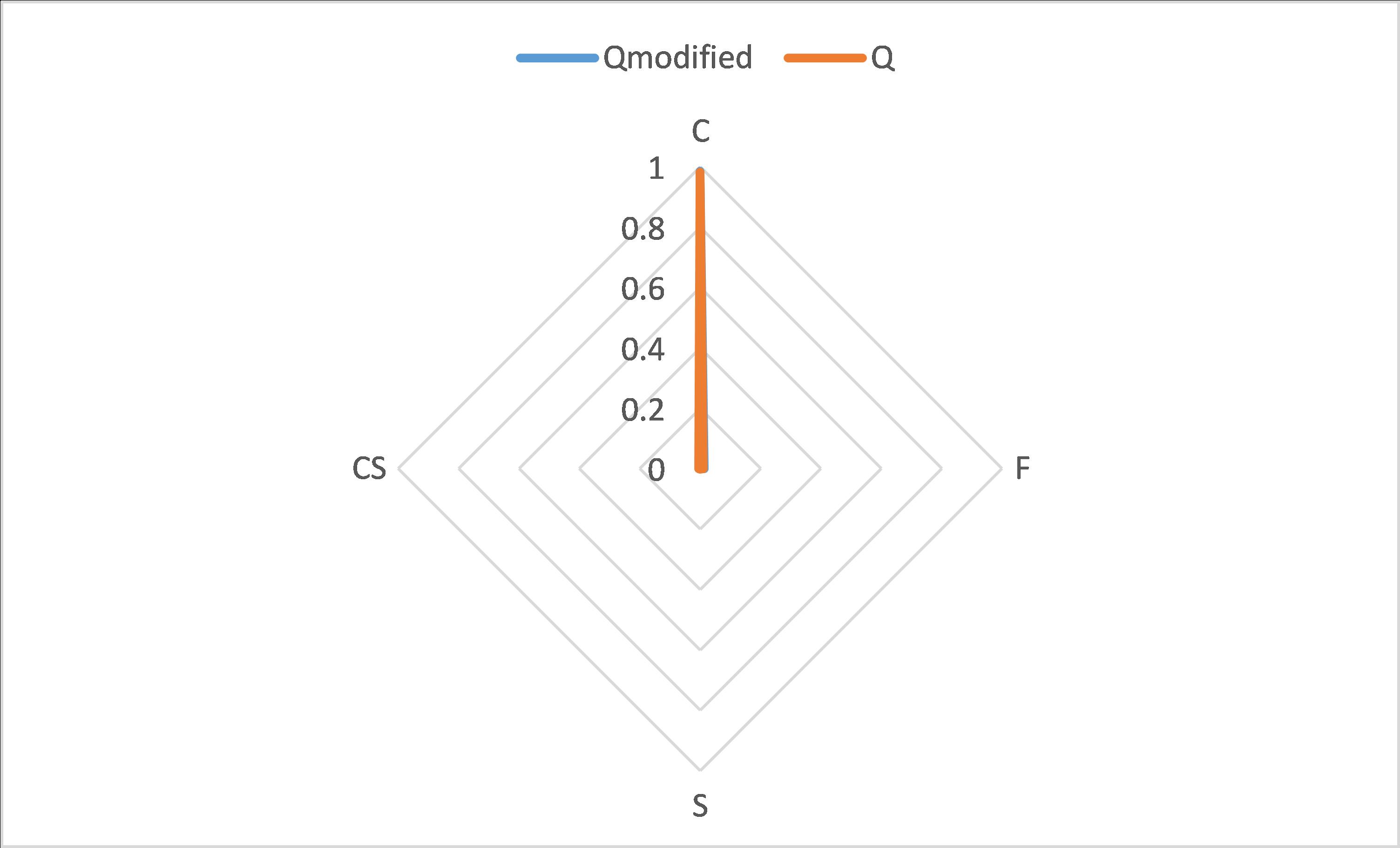} %中括号中的参数是设置图片充满文档的大小，你也可以使用小数来缩小图片的尺寸。
	\caption{The comparison of results combined between proposed method and traditional rule of combination in quantum field in application1} %caption是用来给图片加上图题的
	\label{1} %这是添加标签，方便在文章中引用图片。
\end{figure}
\clearpage

\subsection{Application of decision making}

The system of estimate in quantum scale can not only play an important role in medical diagnosis, but also is powerful in decisions making, which is verified convincingly in actual applications. Under normal circumstances, if the event cannot be observed directly, then researchers are more inclined to introduce related problems into the quantum field to better adapt to such an uncertain environment. The example in the following illustrates the preponderance of the system of judgment presented in this article in decision making.
\begin{table}[h]\footnotesize
	\centering
	\caption{Quantum evidences given by financial experts of application 2}
	\begin{spacing}{1.80}
		\begin{tabular}{c c c c c}\hline
			$Evidences$ &\multicolumn{4}{c}{$Values \ \ of \ \ propositions$}\\\hline
			&$\{AS\}$&$\{BS\}$&$\{ABS\}$&$\{NO\}$\\
			$Evidence_{1}$&$0.8124e^{1.1726j}$&$0.2646e^{1.4496j}$&$0.4359e^{1.5387j}$&$0.2828e^{1.0243j}$\\  
			&$\{BS\}$&$\{AS\}$&$\{ABS\}$&$\{NO\}$\\
			$Evidence_{2}$&$0.7550e^{1.3396j}$&$0.4796e^{0.4907j}$&$0.4123e^{1.2475j}$&$0.1732e^{1.4783j}$\\ 
			&$\{ABS\}$&$\{BS\}$&$\{AS\}$&$\{NO\}$\\
			$Evidence_{3}$&$0.1000e^{1.2451j}$&$0.1732e^{0.4360j}$&$0.9539e^{0.9225j}$&$0.2236e^{1.4317j}$\\ 
			&$\{NO\}$&$\{ABS\}$&$\{BS\}$&$\{AS\}$\\
			$Evidence_{4}$&$0.5196e^{1.5361j}$&$0.200e^{1.3541j}$&$0.5744e^{1.0070j}$&$0.6000e^{1.0720j}$\\ 
			\hline
		\end{tabular}
	\end{spacing}
	%\label{tab:Margin_settings}
	\label{data1}
\end{table}

Assume there exists a financial company which makes decisions about schemes for buying stocks provided by different fiscal experts.  According to the situation of financial market and the forecast of the outlook of the company, diverse stages are given by varying monetary experts about purchasing A-Share, purchasing B-Share, purchasing A-Share and B-Share at the same time and "do nothing". Among them, "do nothing" means that this financial experts suggests the company is not advised to purchase shares of stocks under the conditions of current market.
\begin{table}[h]\footnotesize
	\centering
	\caption{The calculation results parameter $d_{XP}(Q_{i},Q_{j})$ of  application 2}
	\begin{spacing}{1.80}
		\begin{tabular}{c c c c c }\hline
			$Quantum \ \ evidence$ &\multicolumn{4}{c}{ $Values \ \ of \ \ distance \ \ between \ \ QBPAs$} \\\hline
			
			& $AS $& $BS$ & $ABS$ & $NO$  \\ 
			$Evidence_{1}$ &$ 0 $ & $0.1912$  & $0.1773 $ & $0.1656$  \\ 
			$Evidence_{2} $& $0.1912$  & $0$  & $0.1375$  & $0.1728$  \\ 
			$Evidence_{3}$ &$ 0.1773 $ & $0.1375$  & $0$  &$ 0.1556$  \\ 
			$Evidence_{4}$ & $0.1656 $ &$ 0.1728  $&$ 0.1556$  &$ 0 $ \\ \hline
		\end{tabular}
	\end{spacing}
	%\label{tab:Margin_settings}
	\label{XP}
\end{table}

Based on practical situations, the rough figure is drawn preliminarily. The frame of discernment in the form of quantum is defined as $\Theta = \{AS,BS,ABS,NO\}$. $AS$ represents purchasing $A-Share$ and $BS$ denotes buying B-Share. Purchasing A-Share and B-Share at the same time is indicated by $ABS$. "No" represents taking no action in allusion to stocks. Table \ref{data1} demonstrates particular information about given quantum evidence.
\begin{figure}[h] %figure环境，h默认参数是可以浮动，不是固定在当前位置。如果要不浮动，你就可以使用大写float宏包的H参数，固定图片在当前位置，禁止浮动。
	\centering %使图片居中显示
	\includegraphics[width=0.7\textwidth]{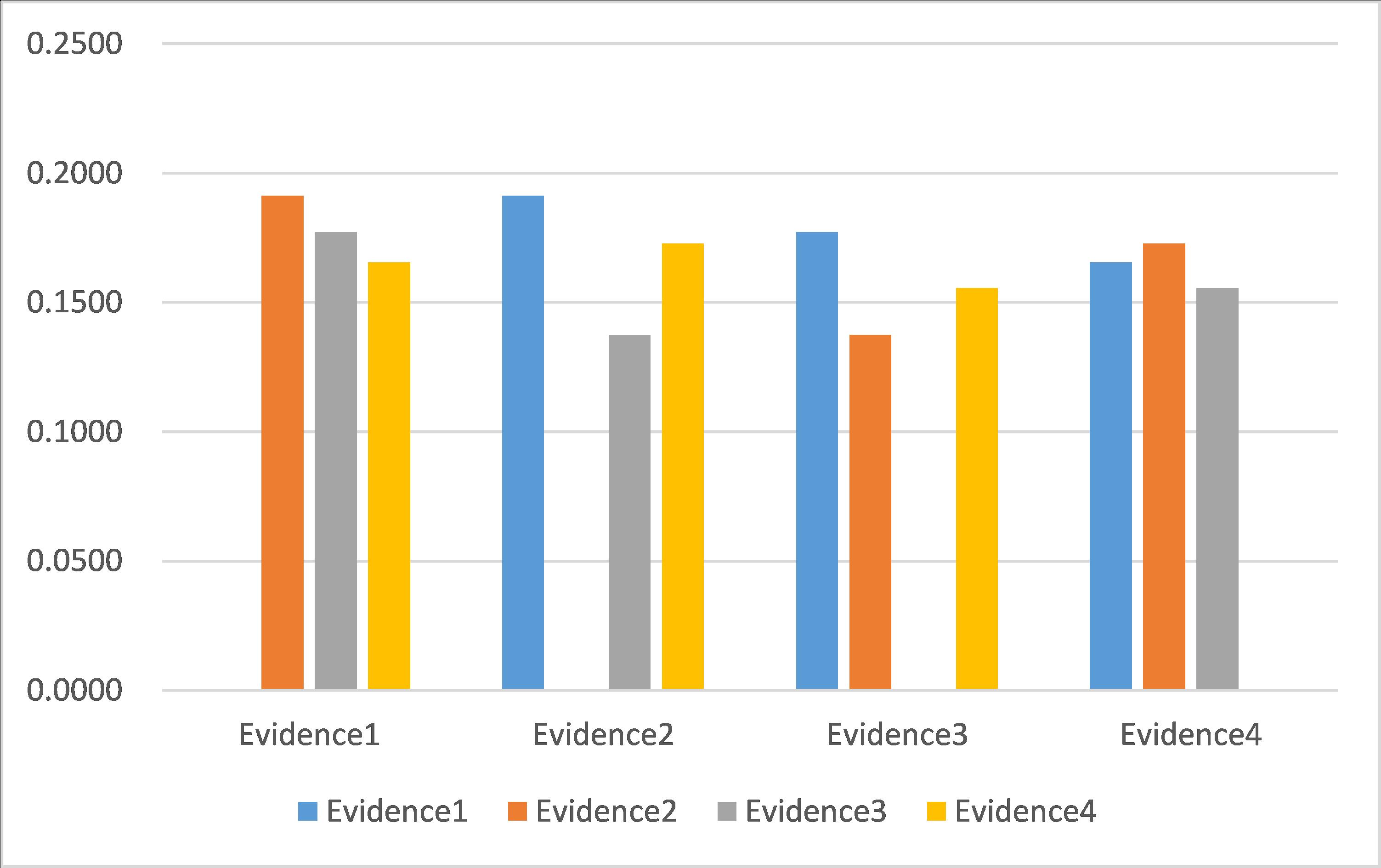} %中括号中的参数是设置图片充满文档的大小，你也可以使用小数来缩小图片的尺寸。
	\caption{The calculation result of parameter $Sim_{2}(Q_{i},Q_{j})$ of application1} %caption是用来给图片加上图题的
	\label{1} %这是添加标签，方便在文章中引用图片。
\end{figure}

Besides, the results of calculation of $d_{XP}(Q_{i},Q_{j})$  and $d_{WB}(Q_{i},Q_{j})$ are listed in Table \ref{XP} and Table \ref{WB}. Then, by utilizing equations of the similarity and two species of discrepancy of quantum evidences, $ Sim_{1}(Q_{i},Q_{j})$ and $ Sim_{2}(Q_{i},Q_{j})$ can be obtained
\begin{table}[h]\footnotesize
	\centering
	\caption{The calculation results parameter $d_{WB}(Q_{i},Q_{j})$ of application 2}
	\begin{spacing}{1.80}
		\begin{tabular}{c c c c c }\hline
			$ Quantum \ \ evidence $ &\multicolumn{4}{c}{ $Values \ \ of \ \ distance \ \ between \ \ QBPAs$ }\\\hline
			
			& $AS $& $BS$ & $ABS$ & $NO$  \\ 
			$Evidence_{1}$ & $0$  &$ 0.2055 $ & $0.2042$  &$ 0.1524$  \\ 
			$Evidence_{2} $ &$0.2055 $ & $0  $&$ 0.1624  $&$ 0.1590$  \\ 
			$Evidence_{3} $ &$0.2042 $ &$ 0.1624 $ &$ 0 $ & $0.1165 $ \\ 
			$Evidence_{4} $ &$0.1524 $ & $0.1590 $ & $0.1165 $ & $0 $ \\ \hline
		\end{tabular}
	\end{spacing}
	%\label{tab:Margin_settings}
	\label{WB}
\end{table}
which are presented in Table \ref{Sim1} and Table \ref{Sim2}. Except for those, factors introduced before are taken account of combination and the weight of provided evidence is enumerated in Table \ref{weights}. In the last, the contrast of consequence which are combined by taking advantage of proposed method and traditional rule of combination in quantum field and at the level of classic probability assignment are listed in Table \ref{quantumvalue} and Table \ref{value} respectively.
\begin{figure}[h] %figure环境，h默认参数是可以浮动，不是固定在当前位置。如果要不浮动，你就可以使用大写float宏包的H参数，固定图片在当前位置，禁止浮动。
	\centering %使图片居中显示
	\includegraphics[width=0.7\textwidth]{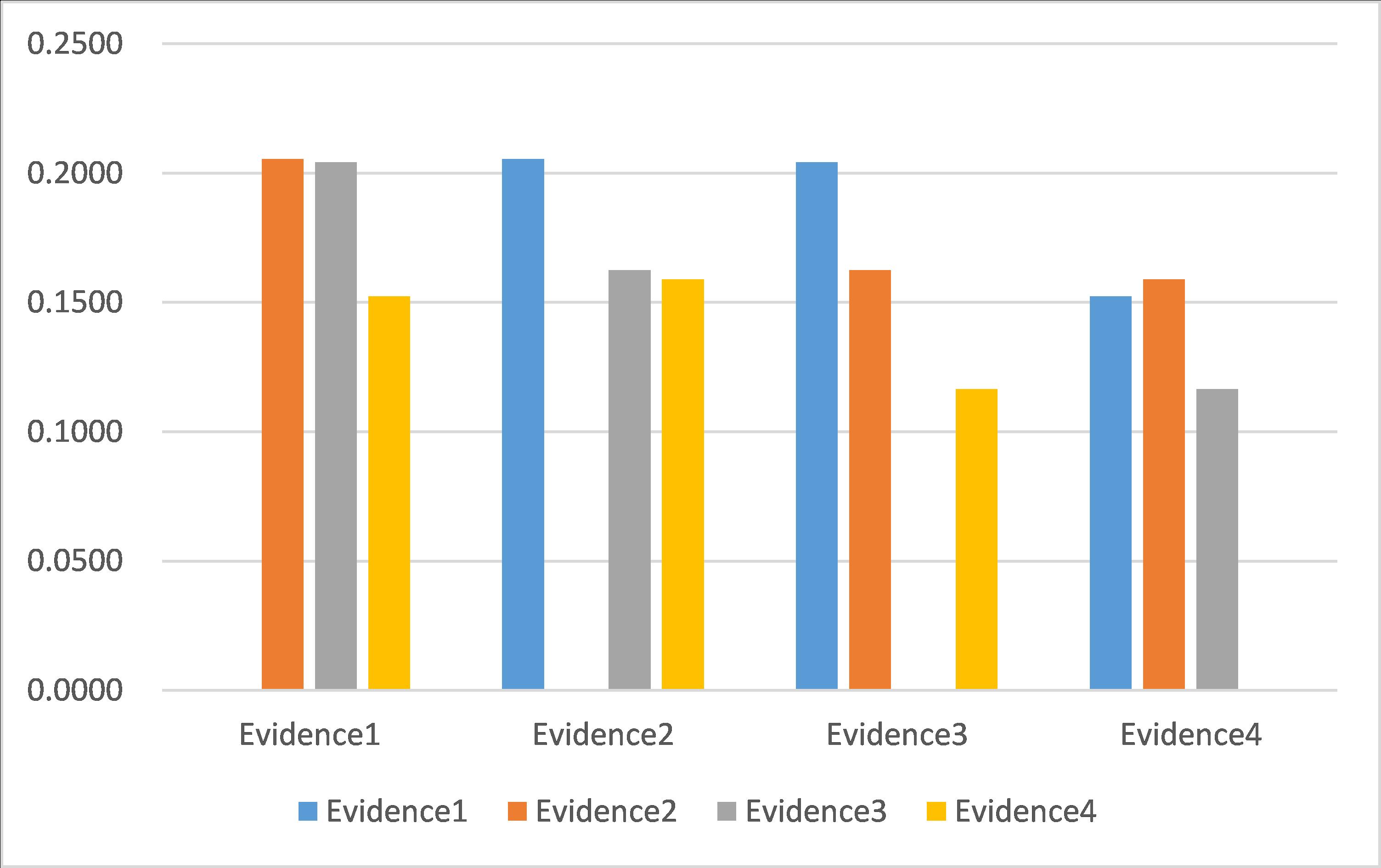} %中括号中的参数是设置图片充满文档的大小，你也可以使用小数来缩小图片的尺寸。
	\caption{The calculation result of parameter $Sim_{2}(Q_{i},Q_{j})$ of application1} %caption是用来给图片加上图题的
	\label{1} %这是添加标签，方便在文章中引用图片。
\end{figure}

\begin{table}[h]\footnotesize
	\centering
	\caption{The calculation results parameter $Sim_{1}(Q_{i},Q_{j})$ of application2}
	\begin{spacing}{1.80}
		\begin{tabular}{c c c c c }\hline
			$Quantum \ \ evidence $ &\multicolumn{4}{c}{ $Values \ \ of \ \ similarity  \ \ between \ \ QBPAs $}\\\hline
			
			& $AS$& $BS$ & $ABS$ & $NO $ \\ 
			$ Evidence_{1} $ & $ 1 $ & $0.0909 $ &$ 0.1484  $& $0.1492$  \\ 
			$ Evidence_{2} $ & $0.0909$  &$ 1  $& 0$.2039 $ & $0.1913 $ \\ 
			$ Evidence_{3}$ &$0.1484 $ &$ 0.2039 $ &$ 1 $ & $0.2163  $\\ 
			$ Evidence_{4} $&$0.1492 $ &$ 0.1913 $ &$ 0.2163 $ &$ 1 $ \\ \hline
		\end{tabular}
	\end{spacing}
	%\label{tab:Margin_settings}
	\label{Sim1}
\end{table}
By analysing the comparison of the traditional method of combination of quantum evidences and the modified approach mentioned above, the proposed method exerts a lesser influence that made the most value decreased on final values combined. From observing the Table \ref{value},purchasing A-Share is confirmed which is
\begin{figure}[h] %figure环境，h默认参数是可以浮动，不是固定在当前位置。如果要不浮动，你就可以使用大写float宏包的H参数，固定图片在当前位置，禁止浮动。
	\centering %使图片居中显示
	\includegraphics[width=0.7\textwidth]{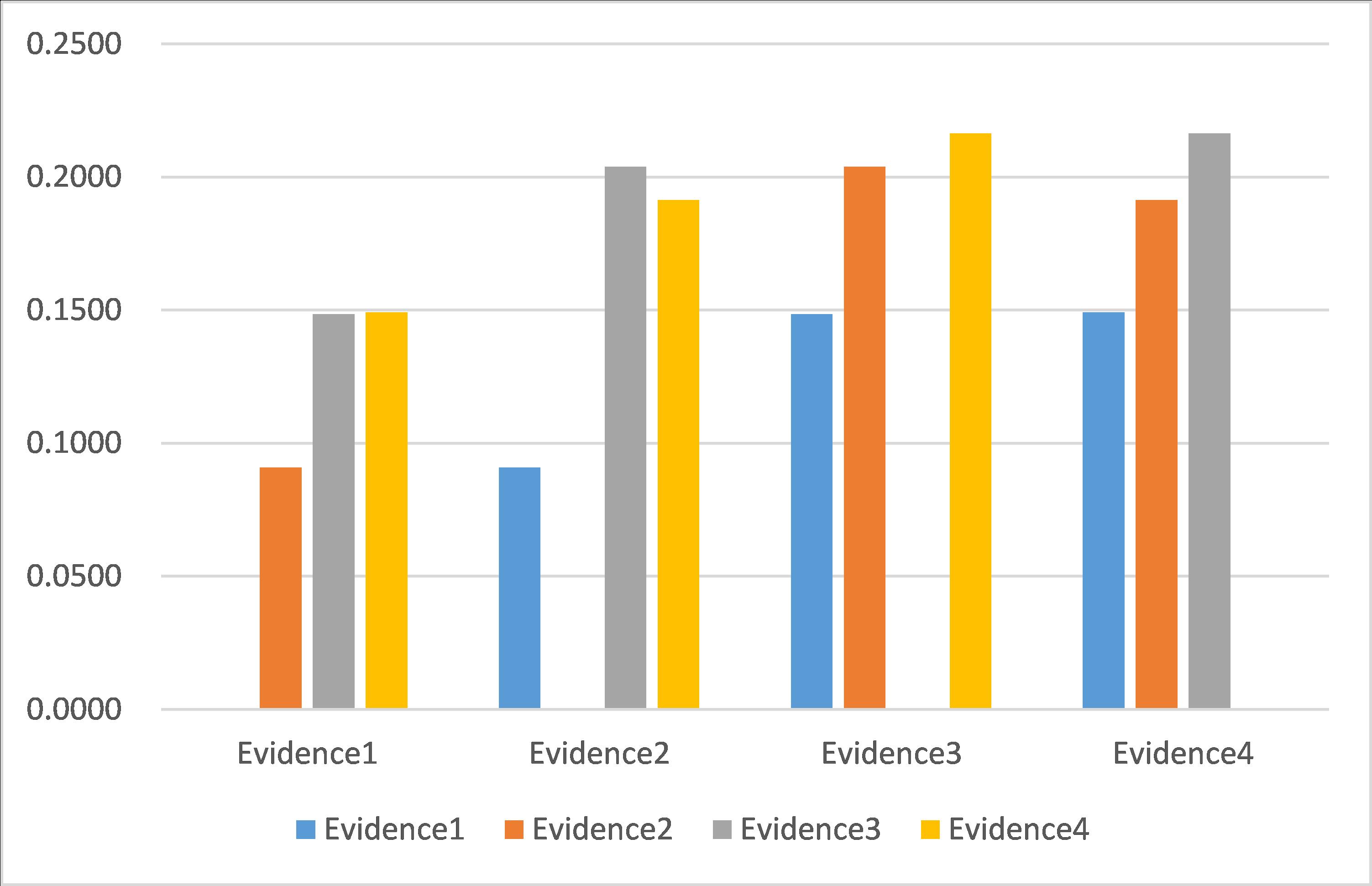} %中括号中的参数是设置图片充满文档的大小，你也可以使用小数来缩小图片的尺寸。
	\caption{The calculation result of parameter $Sim_{2}(Q_{i},Q_{j})$ of application1} %caption是用来给图片加上图题的
	\label{1} %这是添加标签，方便在文章中引用图片。
\end{figure}
the best choice for the company in current environment of market told by two kinds of combined manners. Apparently, the combined values are amended more legitimately by the proposed method mentioned above which made higher value not absolute like traditional method. The modified values accord with reality of life more distinctly and conform to the normal development of things. As a result, purchasing A-Share could obtain revenue maximization. However,
\begin{table}[h]\footnotesize
	\centering
	\caption{The calculation results parameter $Sim_{2}(Q_{i},Q_{j})$ of application 2}
	\begin{spacing}{1.80}
		\begin{tabular}{c c c c c }\hline
			Quantum evidence &\multicolumn{4}{c}{ $Values \ \ of \ \ similarity \ \ between \ \ QBPAs $}\\\hline
			
			& $AS$ & $BS$ & $ABS$ & $NO$  \\ 
			$Evidence_{1} $& $1$  & $0.1541$  & $0.1571 $ &$ 0.1697$  \\ 
			$Evidence_{2} $& $0.1541 $ & $1$  & $0.1733$  & $0.1669 $ \\ 
			$Evidence_{3}$ & $0.1571$  &$ 0.1733$  & $1 $ &$ 0.1790 $ \\ 
			$Evidence_{4}$ & $0.1697 $ &$ 0.1669 $ & $0.1790  $&$ 1$  \\ \hline
		\end{tabular}
	\end{spacing}
	%\label{tab:Margin_settings}
	\label{Sim2}
\end{table}
purchasing B-Share should be taken into consideration at the same time. Because there still exists some certain possibility that buying B-Share benefits the most. More subtle modifications are made by the improved approach without
\begin{figure}[h] %figure环境，h默认参数是可以浮动，不是固定在当前位置。如果要不浮动，你就可以使用大写float宏包的H参数，固定图片在当前位置，禁止浮动。
	\centering %使图片居中显示
	\includegraphics[width=0.7\textwidth]{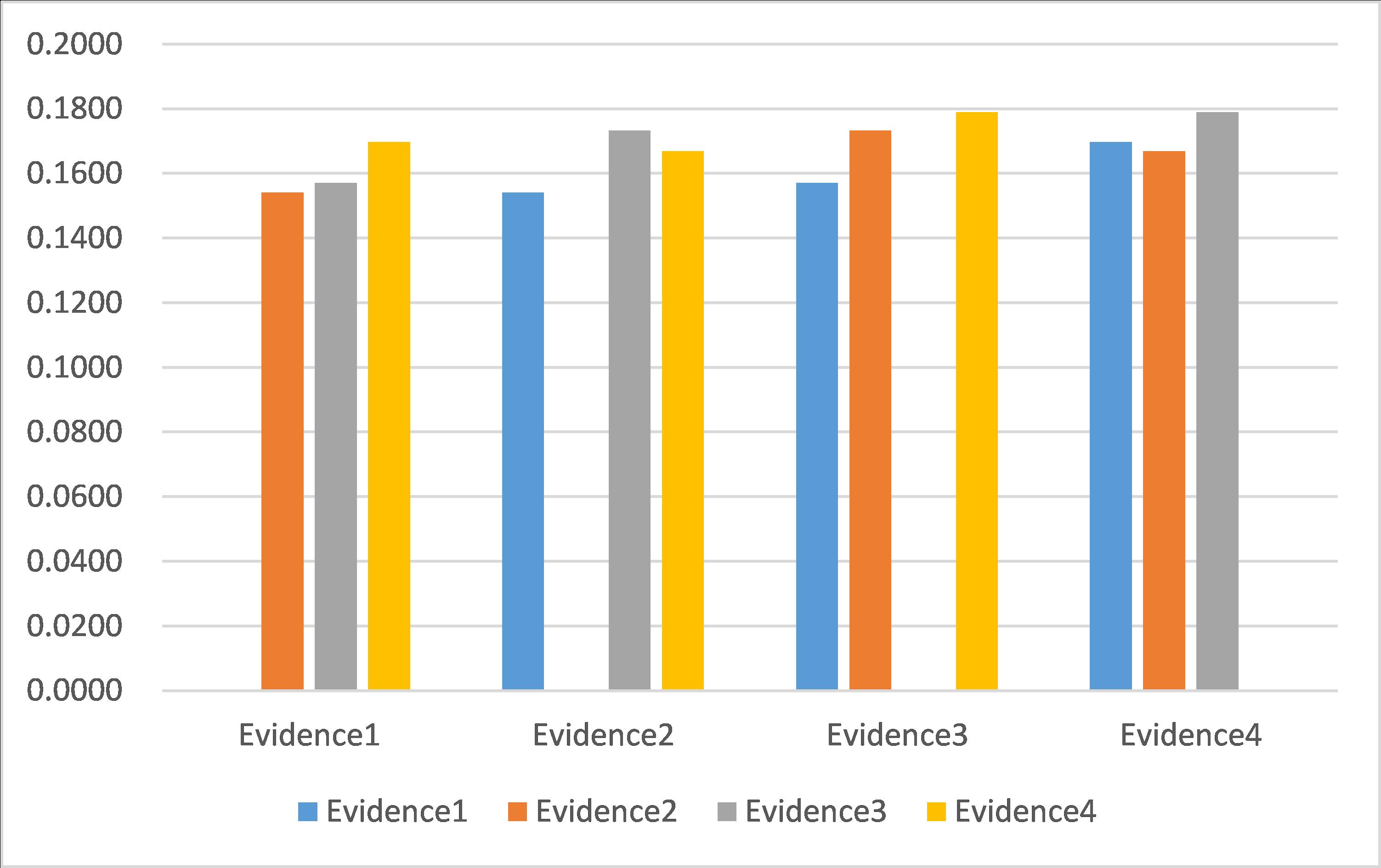} %中括号中的参数是设置图片充满文档的大小，你也可以使用小数来缩小图片的尺寸。
	\caption{The calculation result of parameter $Sim_{2}(Q_{i},Q_{j})$ of application1} %caption是用来给图片加上图题的
	\label{1} %这是添加标签，方便在文章中引用图片。
\end{figure}
changing the original indication. This mainly because the discussion of order between propositions and the measurement of discrepancy cancels out each other which leads to two species of combined results analogical. However, the proposed method is the expansion of traditional combinations which is more forceful and a more veracious combination. Generally speaking, the improved combined values calculated by this method are more accurate and illustrate that the order of propositions in the quantum field is necessary.

\begin{table}[h]\footnotesize
	\centering
	\caption{The calculation results parameter $Wgt_{i}^{nor}$ of application 2}
	\begin{spacing}{1.80}
		\begin{tabular}{c c c c c }\hline
			
			$Quantum \ \ evidence$ & $AS $& $BS$ & $ ABS $ &$ NO $  \\ \hline
			$Wgt_{i}^{nor}$ &$ 0.2173$ & $0.2451 $&$ 0.2695 $& $0.2681 $ \\ \hline
		\end{tabular}
	\end{spacing}
	%\label{tab:Margin_settings}
	\label{weights}
\end{table}

\begin{table}[h]\footnotesize
	\centering
	\caption{The comparison of results combined between proposed method and traditional rule of combination in quantum field in application 2}
	\begin{spacing}{1.80}
		\begin{tabular}{c c c c c }\hline
			
			$Proposition $& $AS $& $BS $& $ABS $& $NO$  \\ \hline
			$The \ \ improved \ \ combined \ \ values$ & $0.8879e^{-2.3298i}$ & $0.4595e^{-1.0344i}$& $0.0129e^{-0.7355i}$ & $0.0150e^{-0.3980i}$ \\ \hline
			$Proposition $ & $AS $& $BS $& $ABS$ & $NO$  \\ \hline
			$Combined \ \ values $ & $0.9325e^{-2.4725i}$& $0.3600e^{-0.9740i}$& $0.0151e^{-0.7444i}$& $0.0237e^{-0.6598i}$  \\ \hline
		\end{tabular}
	\end{spacing}
	%\label{tab:Margin_settings}
	\label{quantumvalue}
\end{table}

\begin{table}[h]\footnotesize
	\centering
	\caption{The comparison of results combined between proposed method and traditional rule of combination in the form of classic probability assignment in application 2}
	\begin{spacing}{1.80}
		\begin{tabular}{c c c c c }\hline
			
			$Proposition $& $AS$ & $BS$ & $ABS$ & $NO$  \\ \hline
			$The \ \ improved \ \ combined \ \ values$ & $0.7885 $&$0.2112 $& $0.0002 $& $0.0002 $  \\ \hline
			$Proposition$ & $AS$ & $BS$ & $ABS$ & $NO $ \\ \hline
			$ Combined \ \ values $ &$ 0.8696 $&$ 0.1296$ &$ 0.0002$ & $0.0006$  \\ \hline
		\end{tabular}
	\end{spacing}
	%\label{tab:Margin_settings}
	\label{value}
\end{table}
\begin{figure}[h] %figure环境，h默认参数是可以浮动，不是固定在当前位置。如果要不浮动，你就可以使用大写float宏包的H参数，固定图片在当前位置，禁止浮动。
	\centering %使图片居中显示
	\includegraphics[width=0.7\textwidth]{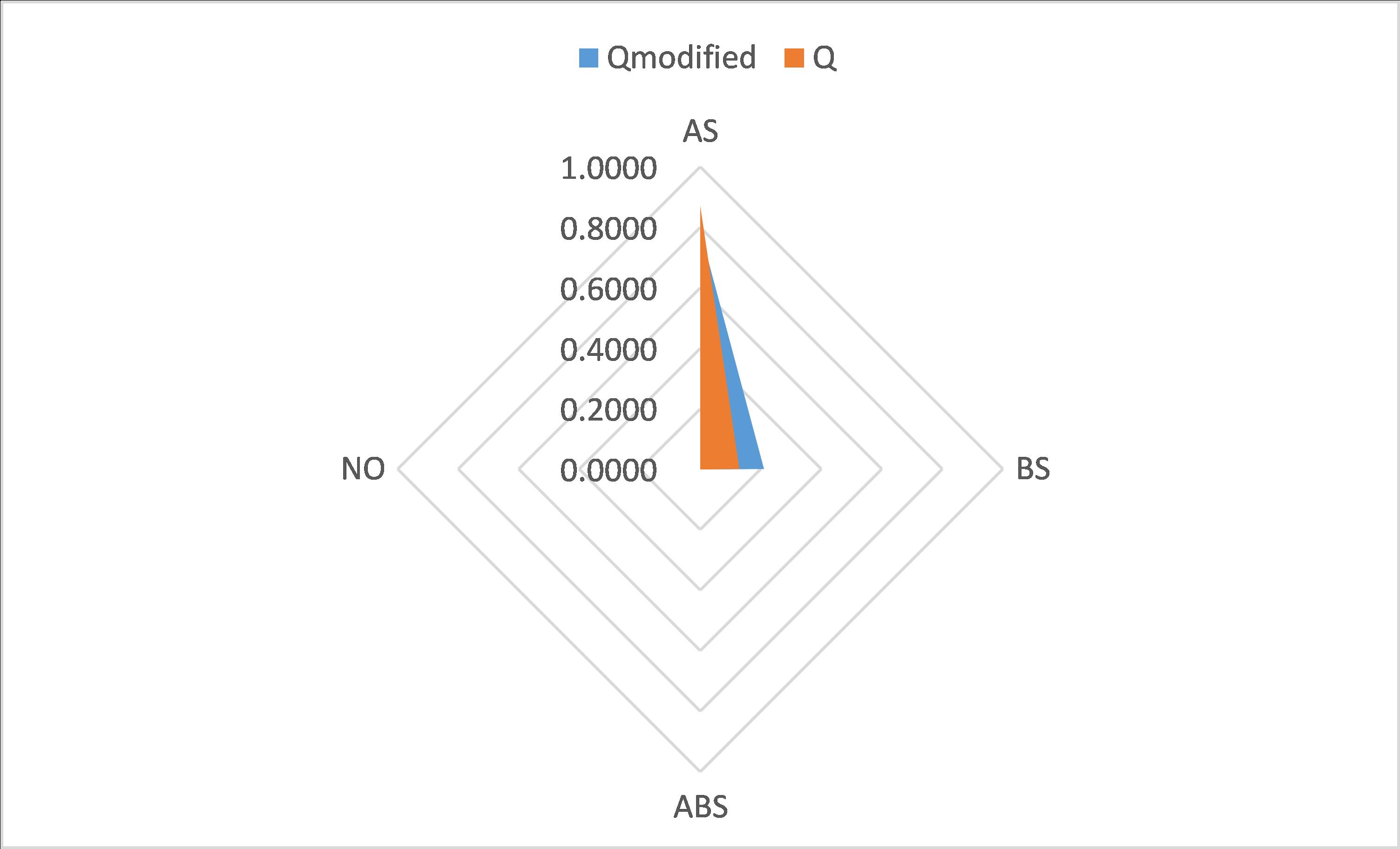} %中括号中的参数是设置图片充满文档的大小，你也可以使用小数来缩小图片的尺寸。
	\caption{The calculation result of parameter $Sim_{2}(Q_{i},Q_{j})$ of application1} %caption是用来给图片加上图题的
	\label{1} %这是添加标签，方便在文章中引用图片。
\end{figure}
\clearpage
\subsection{Application of fault diagnosis}

Suppose that there is a factory equipped with fully automatic instruments. Due to the immature technology, some malfunction may happen on the machine. In order to address the issue, fault detection sensors are designed and attached to the machine. Three parts of the machine is possible to break down and the information of faults is given in the form of quantum information.
\begin{table}[h]\footnotesize\
	\centering
	\caption{Evidences given by fault detection sensors}
	\begin{spacing}{1.80}
		\begin{tabular}{c c c c c c}\hline
			$Evidences$ & \multicolumn{5}{c}{$Values \ \ of \ \ propositions$}\\\hline
			&$\{S\}$&$\{M\}$&$\{E\}$&$\{SM\}$&$\{ME\}$\\
			$Evidence_{1}$&$0.5000e^{1.1196j}$&$0.3874e^{0.5798j}$&$0.5830e^{1.3477j}$&$0.3317e^{1.2866j}$&$0.3873e^{1.4491j}$\\  
			&$\{E\}$&$\{SM\}$&$\{ME\}$&$\{M\}$&$\{S\}$\\
			$Evidence_{2}$&$0.4796e^{1.1702j}$&$0.4123e^{1.4809j}$&$0.3873e^{1.2312j}$&$0.4583e^{1.1719j}$&$0.4899e^{1.3296j}$\\ 
			&$\{SM\}$&$\{S\}$&$\{M\}$&$\{ME\}$&$\{E\}$\\
			$Evidence_{3}$&$0.3742e^{1.2330j}$&$0.4123e^{1.3032j}$&$0.4359e^{0.7049j}$&$0.4899^{1.5075j}$&$0.5099e^{1.3473j}$\\ 	
			&$\{ME\}$&$\{E\}$&$\{S\}$&$\{M\}$&$\{SM\}$\\
			$Evidence_{4}$&$0.4243e^{1.4360j}$&$0.5568e^{1.1665j}$&$0.4796e^{1.2017j}$&$0.4123^{0.0798j}$&$0.3317e^{1.4681j}$\\ 
			&$\{M\}$&$\{ME\}$&$\{E\}$&$\{S\}$&$\{SM\}$\\
			$Evidence_{5}$&$0.5196e^{0.8901j}$&$0.4472e^{1.4969j}$&$0.4123e^{1.3183j}$&$0.4359e^{0.5483j}$&$0.4123e^{1.5069j}$\\ 
			\hline
		\end{tabular}
	\end{spacing}
	%\label{tab:Margin_settings}
	\label{w2}
\end{table}

Based on the situations mentioned above, the quantum frames of discernment of the specific problems is defined as $\Theta=\{S,M,E,FM,ME\}$. $S$ denotes that the fault which happens in front part and $M$ presents that the middle part breaks down . In addition, $E$ indicates fault at end part and $SM$ means the failure occur in both the beginning and the middle part. What's more, the failures occur in both of the middle part and the end is presented by $ME$. Evidences for the fault diagnosis is given in Table\ref{w2}. Propositions have different orders in different kinds of evidence and the more advanced position they are means the higher weight they have.

\begin{table}[h]\footnotesize
	\centering
	\caption{The calculation result of parameter $d_{XP}(Q_{i},Q_{j})$ of application3}
	\begin{spacing}{1.80}
		\begin{tabular}{c c c c c c }\hline
			$Quantum \ \ evidence $ & \multicolumn{5}{c}{Values of distance between QBPAs}\\\hline
			
			& $ Evidence_{1} $ & $ Evidence_{2} $ & $ Evidence_{3} $ & $ Evidence_{4} $ & $ Evidence_{5}$ \\
			$Evidence_{1} $ & $   0 $ & $ 0.095222009 $ & $ 0.061335424 $ & $ 0.128291791 $ & $ 0.08803936$ \\
			$Evidence_{2} $ & $ 0.095222009 $ & $ 0 $ & $ 0.065130772 $ & $ 0.072923152 $ & $ 0.092479968 $\\
			$Evidence_{3} $ & $ 0.061335424 $ & $ 0.065130772 $ & $ 0 $ & $ 0.09135496 $ & $ 0.18540361$ \\
			$Evidence_{4} $ & $ 0.128291791 $ & $ 0.072923152 $ & $ 0.09135496 $ & $ 0 $ & $ 0.119818954$ \\
			$Evidence_{5} $ & $ 0.08803936 $ & $ 0.092479968 $ & $ 0.18540361 $ & $ 0.119818954 $ & $ 0$ \\ \hline
		\end{tabular}
	\end{spacing}
	%\label{tab:Margin_settings}
	\label{w2XP}
\end{table}

The above data couldn't tell which part of the machine fails obviously, to Further determine the location of the fault, the data is processed by several ways. Table \ref{w2XP} and Table \ref{w2WB} show the computed result of $d_{XP}(Q_{i},Q_{j})$  and $d_{WB}(Q_{i},Q_{j})$ respectively. Then, Table \ref{w2Sim1} and Table \ref{w2Sim2} present $ Sim_{1}(Q_{i},Q_{j})$ and $
 Sim_{2}(Q_{i},Q_{j}) $ calculated by applying regulations of the similarity
 \begin{figure}[h] %figure环境，h默认参数是可以浮动，不是固定在当前位置。如果要不浮动，你就可以使用大写float宏包的H参数，固定图片在当前位置，禁止浮动。
 	\centering %使图片居中显示
 	\includegraphics[width=0.7\textwidth]{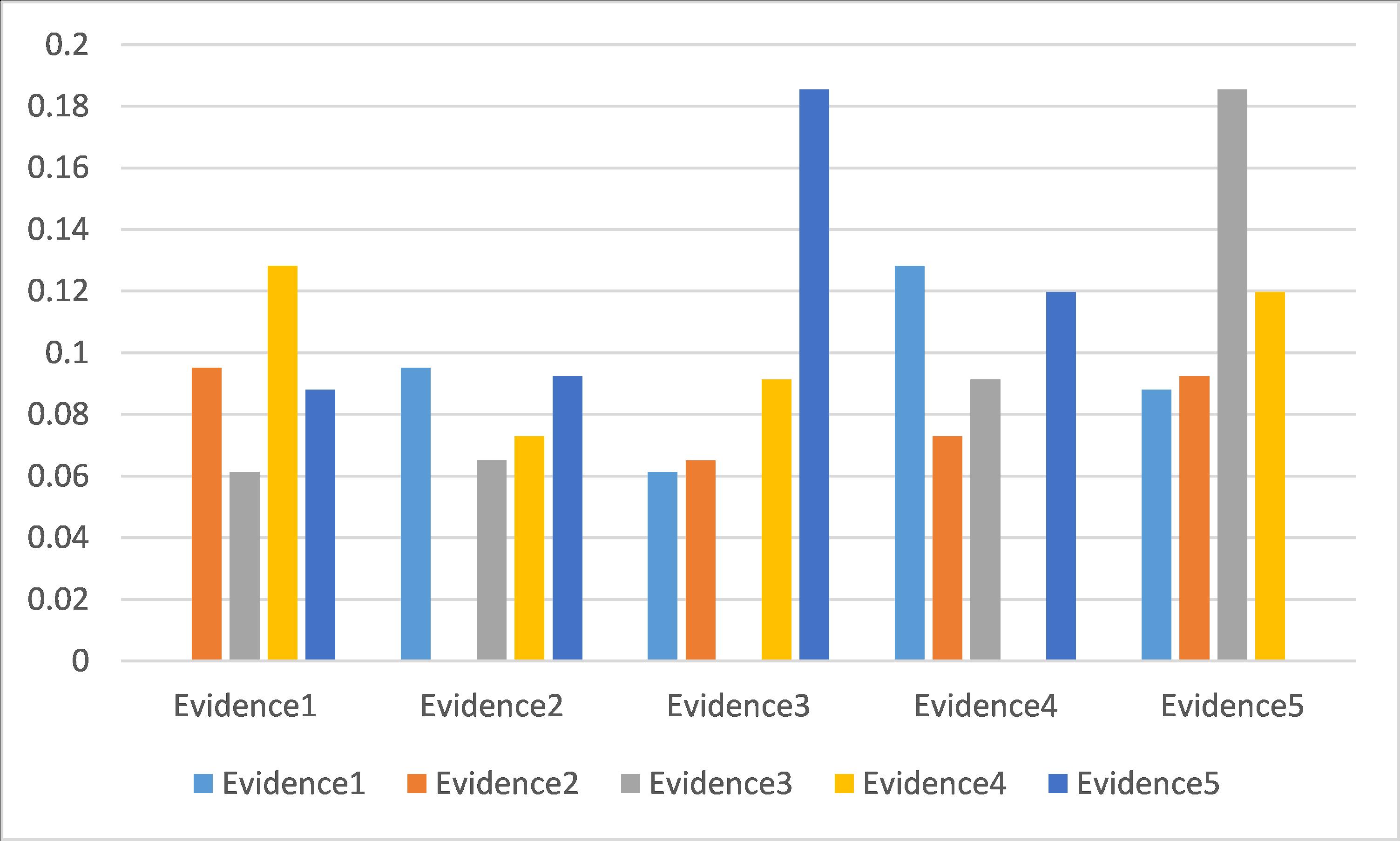} %中括号中的参数是设置图片充满文档的大小，你也可以使用小数来缩小图片的尺寸。
 	\caption{The calculation result of parameter $d_{XP}(Q_{i},Q_{j})$ of application3} %caption是用来给图片加上图题的
 	\label{1} %这是添加标签，方便在文章中引用图片。
 \end{figure}
and two kinds of divergences among quantum evidences. Moreover, to show the significant degree for each evidence, two types of similarity discussed before are synthesized and listed in Table \ref{w2weights}. In the last, Table \ref{w2quantumvalue} and Table \ref{w2value} show the comparison of outcomes which are combined by proposed method and traditional rule of composition respectively.

\begin{table}[h]\footnotesize
	\centering
	\caption{The calculation result of parameter $d_{WB}(Q_{i},Q_{j})$ of application3}
	\begin{spacing}{1.80}
		\begin{tabular}{c c c c c c }\hline
			$			Quantum\ \ evidence $ & \multicolumn{5}{c}{Values of distance between QBPAs}\\\hline
			
			& $ Evidence_{1} $ & $ Evidence_{2} $ & $ Evidence_{3} $ & $ Evidence_{4} $ & $ Evidence_{5}$ \\
			$Evidence_{1} $ & $ 0 $ & $ 0.083689353 $ & $ 0.026769203 $ & $ 0.043637639 $ & $ 0.108407669$ \\
			$	Evidence_{2} $ & $ 0.083689353 $ & $ 0 $ & $ 0.066969002 $ & $ 0.181357579 $ & $ 0.122602905$ \\
			$	Evidence_{3} $ & $ 0.026769203 $ & $ 0.066969002 $ & $ 0 $ & $ 0.074733965 $ & $ 0.103655864$ \\
			$	Evidence_{4} $ & $ 0.043637639 $ & $ 0.181357579 $ & $ 0.074733965 $ & $ 0 $ & $ 0.188176822$ \\
			$	Evidence_{5} $ & $ 0.108407669 $ & $ 0.122602905 $ & $ 0.103655864 $ & $ 0.188176822 $ & $ 0$ \\ \hline
		\end{tabular}
	\end{spacing}
	%\label{tab:Margin_settings}
	\label{w2WB}
\end{table}

It’s obviously concluded in the table \ref{w2value} that outcomes of combination of modified evidences are totally different with the ones calculated by raw evidences. It's believed that each proposition is equally likely to occur by the evidence obtained by combination of raw data while the modified evidences indicates that the end part of the machine break down. Meanwhile, The
\begin{figure}[h] %figure环境，h默认参数是可以浮动，不是固定在当前位置。如果要不浮动，你就可以使用大写float宏包的H参数，固定图片在当前位置，禁止浮动。
	\centering %使图片居中显示
	\includegraphics[width=0.7\textwidth]{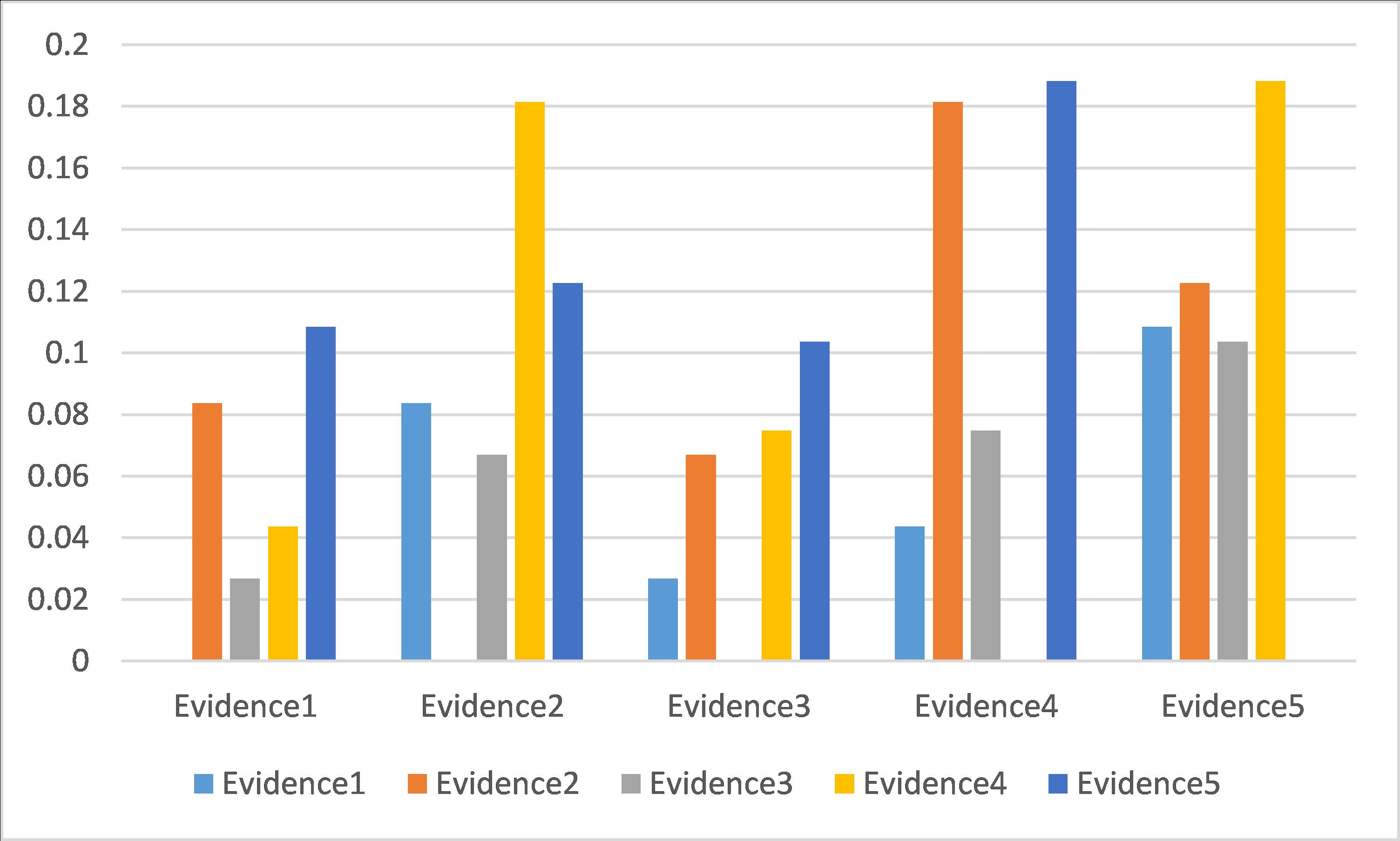} %中括号中的参数是设置图片充满文档的大小，你也可以使用小数来缩小图片的尺寸。
	\caption{The calculation result of parameter $d_{WB}(Q_{i},Q_{j})$ of application3} %caption是用来给图片加上图题的
	\label{1} %这是添加标签，方便在文章中引用图片。
\end{figure}
\begin{table}[h]\footnotesize
	\centering
	\caption{The calculation result of parameter $Sim_{1}$ of application3}
	\begin{spacing}{1.80}
		\begin{tabular}{c c c c c c }\hline
			$Quantum\ \ evidence $ & \multicolumn{5}{c}{Values of similarity between QBPAs}\\\hline
			
			& $ Evidence_{1} $ & $ Evidence_{2} $ & $ Evidence_{3} $ & $ Evidence_{4} $ & $ Evidence_{5} $\\
			$Evidence_{1} $ & $   1 $ & $ 0.096500067 $ & $ 0.12415855 $ & $ 0.104560777 $ & $ 0.100928249$ \\
			$Evidence_{2} $ & $ 0.096500067 $ & $ 1 $ & $ 0.096621301 $ & $ 0.103247208 $ & $ 0.095194423$ \\
			$Evidence_{3} $ & $ 0.12415855 $ & $ 0.096621301 $ & $ 1 $ & $ 0.085874652 $ & $ 0.106825599$ \\
			$Evidence_{4} $ & $ 0.104560777 $ & $ 0.103247208 $ & $ 0.085874652 $ & $ 1 $ & $ 0.086089173$ \\
			$	Evidence_{5} $ & $ 0.100928249 $ & $ 0.095194423 $ & $ 0.106825599 $ & $ 0.086089173 $ & $ 1$ \\ \hline
		\end{tabular}
	\end{spacing}
	%\label{tab:Margin_settings}
	\label{w2Sim1}
\end{table}
probability of proposition $SM$  occurring in the synthetic results of the proposed method is very low. It illustrates that the order of propositions plays a key role in the process of combination. In the fault diagnosis, when the end part fails, the possibility of other parts' failure decreases
\begin{figure}[h] %figure环境，h默认参数是可以浮动，不是固定在当前位置。如果要不浮动，你就可以使用大写float宏包的H参数，固定图片在当前位置，禁止浮动。
	\centering %使图片居中显示
	\includegraphics[width=0.7\textwidth]{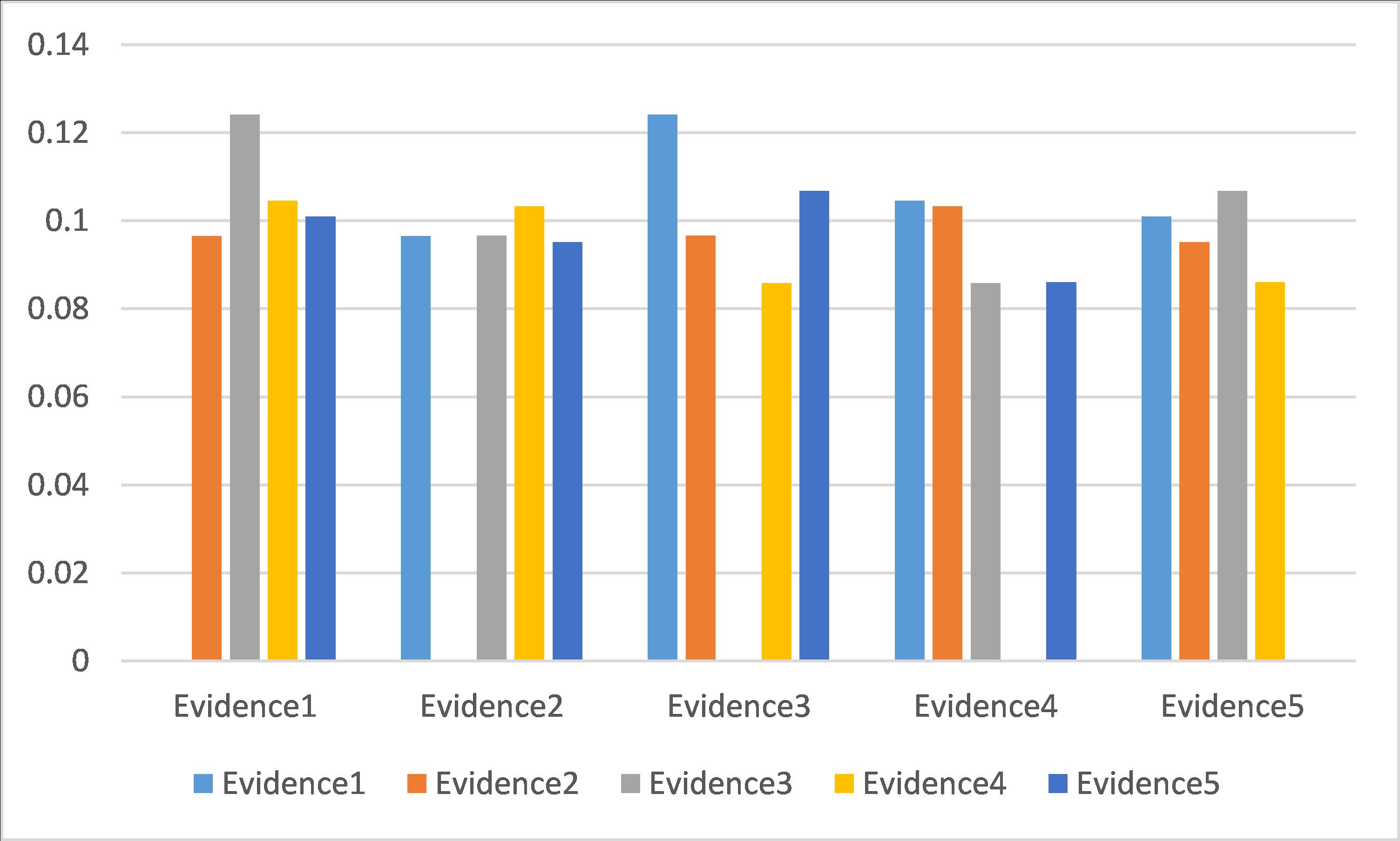} %中括号中的参数是设置图片充满文档的大小，你也可以使用小数来缩小图片的尺寸。
	\caption{The calculation result of parameter $Sim_{1}$ of application 3} %caption是用来给图片加上图题的
	\label{1} %这是添加标签，方便在文章中引用图片。
\end{figure}
accordingly actually. So, proposition $E$ is supposed to assigned higher weights to modify the output of combination and proposition $ME$ is assigned lower weights. The proposed method gives higher weight to the proposition listed more advanced. As thus, The outputs acquired is the interaction of the order of propositions and classic probability assignment which is of great significance for process evidences that.

\begin{table}[h]\footnotesize
	\centering
	\caption{The calculation result of parameter $Sim_{2}$ of application 3}
	\begin{spacing}{1.80}
		\begin{tabular}{c c c c c c }\hline
			$Quantum\ \ evidence $ & \multicolumn{5}{c}{Values of similarity between QBPAs}\\\hline
			
			& $ Evidence_{1} $ & $ Evidence_{2} $ & $ Evidence_{3} $ & $ Evidence_{4} $ & $ Evidence_{5}$ \\
			$Evidence_{1} $ & $    1 $ & $ 0.103190835 $ & $ 0.113738589 $ & $ 0.111304902 $ & $ 0.098754391$ \\
			$Evidence_{2} $ & $ 0.103190835 $ & $ 1 $ & $ 0.105804526 $ & $ 0.091261514 $ & $ 0.098198746$ \\
			$Evidence_{3} $ & $ 0.113738589 $ & $ 0.105804526 $ & $ 1 $ & $ 0.105285393 $ & $ 0.084351484$ \\
			$Evidence_{4} $ & $ 0.111304902 $ & $ 0.091261514 $ & $ 0.105285393 $ & $ 1 $ & $ 0.088109621$ \\
			$Evidence_{5} $ & $ 0.098754391 $ & $ 0.098198746 $ & $ 0.084351484 $ & $ 0.088109621 $ & $ 1$ \\ \hline
		\end{tabular}
	\end{spacing}
	%\label{tab:Margin_settings}
	\label{w2Sim2}
\end{table}
\begin{figure}[h] %figure环境，h默认参数是可以浮动，不是固定在当前位置。如果要不浮动，你就可以使用大写float宏包的H参数，固定图片在当前位置，禁止浮动。
	\centering %使图片居中显示
	\includegraphics[width=0.7\textwidth]{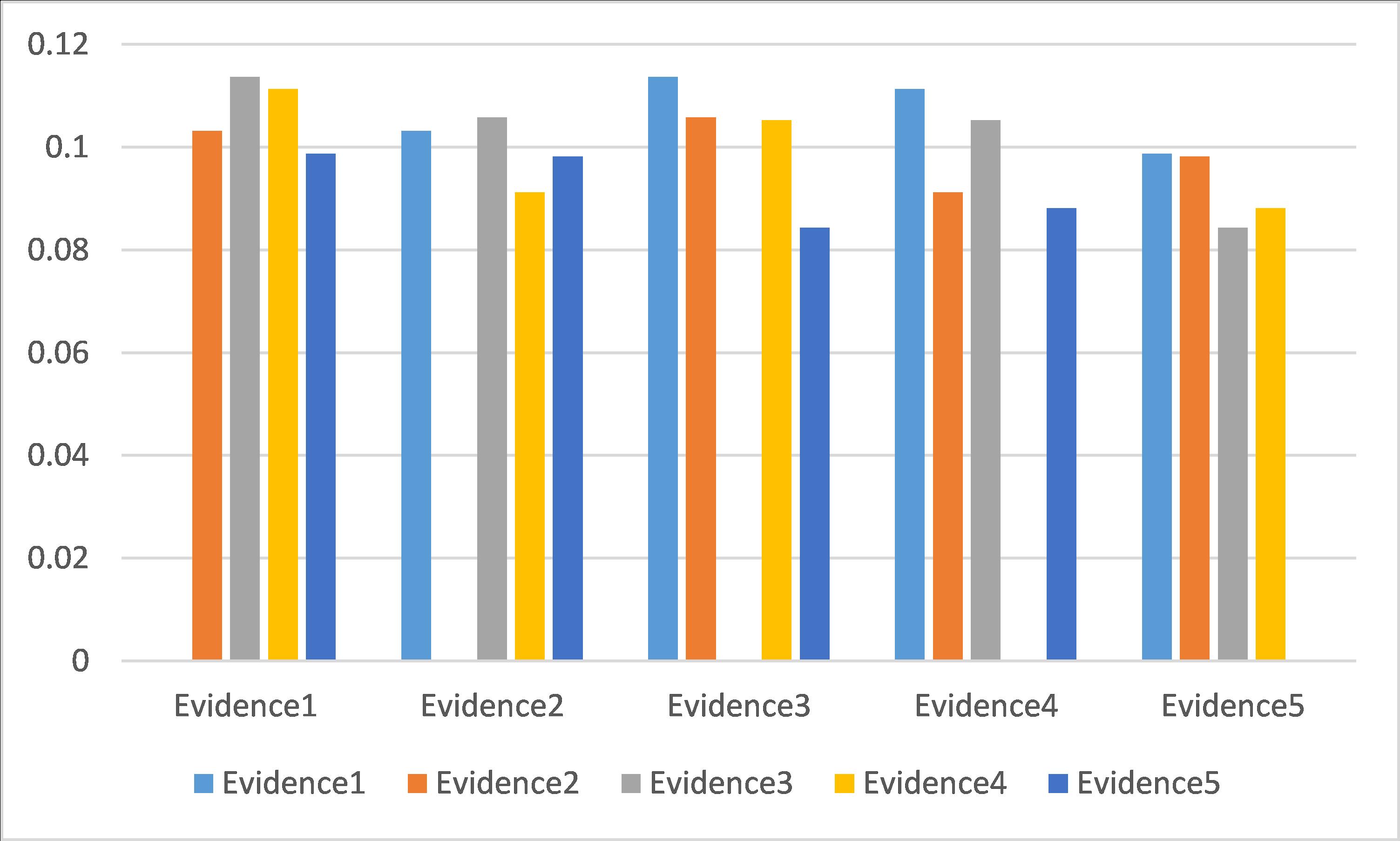} %中括号中的参数是设置图片充满文档的大小，你也可以使用小数来缩小图片的尺寸。
	\caption{The calculation result of parameter $Sim_{2}$ of application 3} %caption是用来给图片加上图题的
	\label{1} %这是添加标签，方便在文章中引用图片。
\end{figure}

\begin{table}[h]\footnotesize
	\centering
	\caption{The calculation result of parameter $Wgt_{i}^{nor}$ of application 3}
	\begin{spacing}{1.80}
		\begin{tabular}{c c c c c c }\hline
			
			$Quantum\ \ evidence $ & $ Evidence_{1} $ & $ Evidence_{2} $ & $ Evidence_{3} $ & $ Evidence_{4} $ & $ Evidence_{5}$ \\ \hline
			$WEIGHT(NOR) $ & $ 0.2133 $ & $ 0.1975 $ & $ 0.2057 $ & $ 0.1939 $ & $ 0.1896$ \\ \hline
		\end{tabular}
	\end{spacing}
	%\label{tab:Margin_settings}
	\label{w2weights}
\end{table}

\begin{table}[h]\footnotesize
	\centering
	\caption{The comparison of results combined between proposed method and traditional rule of combination in quantum field in application 3}
	\begin{spacing}{1.80}
		\begin{tabular}{c c c c c c}\hline
			
			$Proposition $ & $ M $ & $ S $ & $ E $ & $ SM $ & $ ME$ \\ \hline
			$Values\ \ of\ \ combination $ & $0.3540e^{-2.3177i}$ & $0.3699e^{-0.3635i}$ & $0.8053e^{0.3157i}$ & $0.0994e^{0.9264i}$ & $0.2819e^{1.1772i}$\\ \hline
			$Proposition $ & $ M $ & $ S $ & $ E $ & $ SM $ & $ ME$ \\ \hline
			$Values\ \ of\ \ basic\ \ combination $ & $0.3985e^{0.1717i}$ & $0.4724e^{0.7727i}$ & $0.4615e^{0.4848i}$ & $0.4256e^{-0,2590i}$   & $0.4731e^{0.1501i}$\\ \hline
		\end{tabular}
	\end{spacing}
	%\label{tab:Margin_settings}
	\label{w2quantumvalue}
	
\end{table}

\begin{table}[h]\footnotesize
	\centering
	\caption{The comparison of results combined between proposed method and traditional rule of combination in quantum field in application 3}
	\begin{spacing}{1.80}
		\begin{tabular}{c c c c c c}\hline
			
			$Proposition $ & $ M $ & $ S $ & $ E $ & $ SM $ & $ ME$ \\ \hline
			$Values\ \ of\ \ modified\ \ combination $ & $ 0.1253 $ & $ 0.1368 $ & $ 0.6485 $ & $ 0.0099 $ & $ 0.0794 $  \\ \hline
			$Proposition $ & $ M $ & $ S $ & $ E $ & $ SM $ & $ ME$ \\ \hline
			$Values \ \ of\ \ basic\ \ combination $ & $ 0.1588 $ & $ 0.2231 $ & $ 0.2130 $ & $ 0.1812 $ & $ 0.2238 $  \\ \hline
		\end{tabular}
	\end{spacing}
	%\label{tab:Margin_settings}
	\label{w2value}
\end{table}
\begin{figure}[h] %figure环境，h默认参数是可以浮动，不是固定在当前位置。如果要不浮动，你就可以使用大写float宏包的H参数，固定图片在当前位置，禁止浮动。
	\centering %使图片居中显示
	\includegraphics[width=0.7\textwidth]{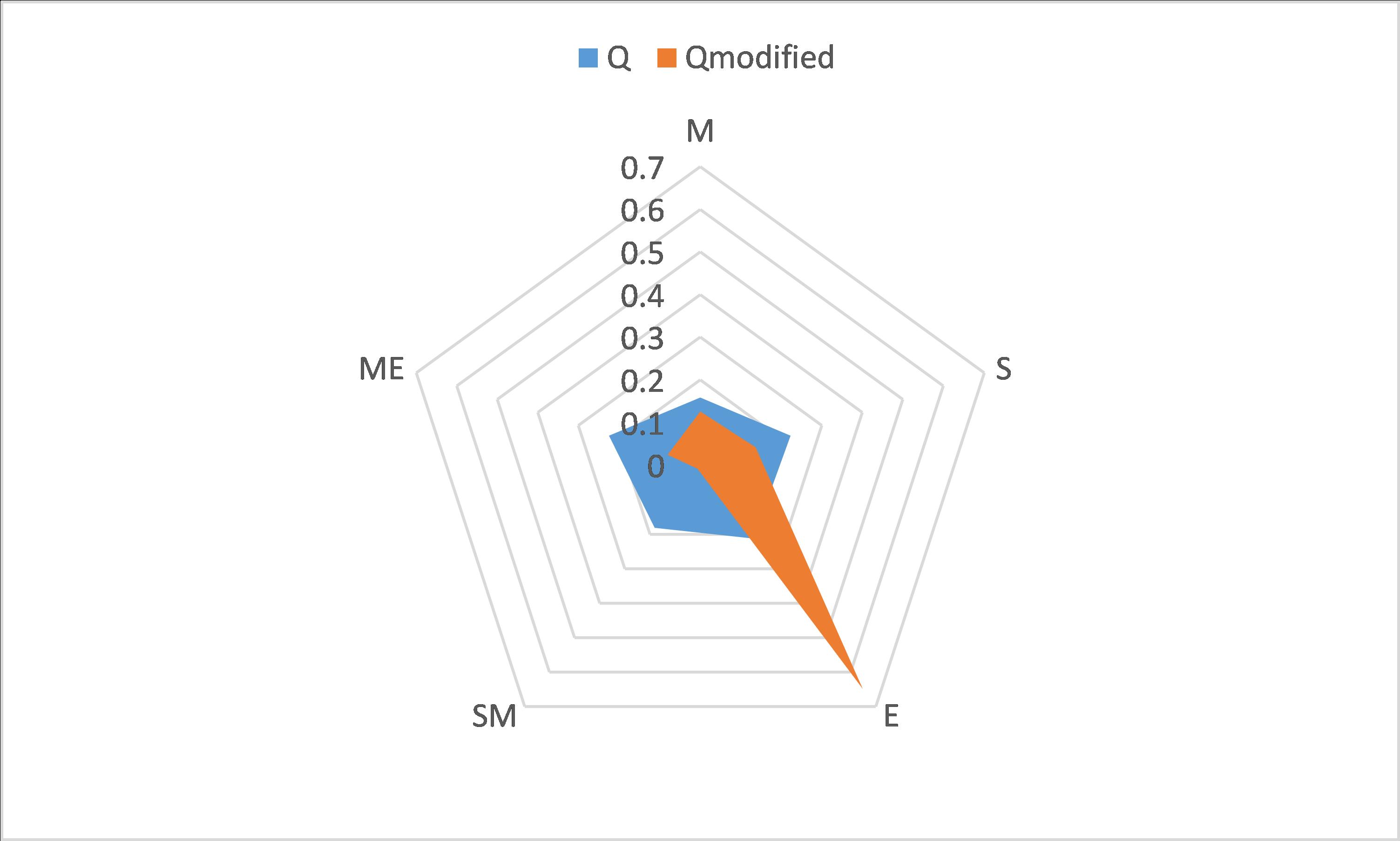} %中括号中的参数是设置图片充满文档的大小，你也可以使用小数来缩小图片的尺寸。
	\caption{The comparison of results combined between proposed method and traditional rule of combination in quantum field in application3} %caption是用来给图片加上图题的
	\label{1} %这是添加标签，方便在文章中引用图片。
\end{figure}

\clearpage

\subsection{Application of income estimate}

The system of judgment also has wide space for the development of application in aspect of work of true matter. In the practical application in regard to life, the ordinal frame of quantum discernment plays an important role in the application of prospect prediction. The example in the following illustrates the preponderance of the system of decision making presented in income estimation.
\begin{table}[h]\footnotesize\
	\centering
	\caption{Quantum evidences given by financial experts of application 4}
	\begin{spacing}{1.80}
		\begin{tabular}{c c c c c c}\hline
			$Quantum \ \ evidences$ &\multicolumn{5}{c}{$Values \ \ of \ \ propositions$}\\\hline
			&$\{Fir\}$&$\{Sec\}$&$\{Thi\}$&$\{Fif\}$&$\{Fou\}$\\
			$Evidence_{SM}$&$0.5568e^{1.3462j}$&$0.5916e^{0.2446j}$&$0.3316e^{1.4590j}$&$0.3606e^{1.5181j}$&$0.3162e^{1.3896j}$\\  
			&$\{Sec\}$&$\{Fir\}$&$\{Fou\}$&$\{Fif\}$&$\{Thi\}$\\
			$Evidence_{RD}$&$0.300e^{1.4639j}$&$0.4123e^{1.5417j}$&$0.5831e^{1.2588j}$&$0.5100e^{0.3383j}$&$0.3741e^{1.1815j}$\\ 
			&$\{Thi\}$&$\{Sec\}$&$\{Fir\}$&$\{Fif\}$&$\{Fou\}$\\
			$Evidence_{ProD}$&$0.3742e^{1.5120j}$&$0.5385e^{1.1540j}$&$0.4359^{0.6650j}$&$0.4583e^{1.5402j}$&$0.4123e^{1.5441j}$\\ 
			&$\{Thi\}$&$\{Sec\}$&$\{Fou\}$&$\{Fir\}$&$\{Fif\}$\\
			$Evidence_{AD}$&$0.3606e^{1.4764j}$&$0.6245e^{0.5603j}$&$0.4796e^{0.9135j}$&$0.400e^{1.5383j}$&$0.300e^{1.4807j}$\\ 
			&$\{Fou\}$&$\{Fif\}$&$\{Fir\}$&$\{Sec\}$&$\{Thi\}$\\
			$Evidence_{PerD}$&$0.5196e^{0.2961j}$&$0.4123e^{1.0891j}$&$0.4359e^{1.3086j}$&$0.4123e^{1.5344j}$&$0.5196e^{0.2961j}$\\ 
			\hline
		\end{tabular}
	\end{spacing}
	%\label{tab:Margin_settings}
	\label{data2}
\end{table}

Assume a company which invites the leaders of its five major functional departments to put forward  income estimates of the company at the end of year for the coming year who come from five core functional departments: Sales \& Marketing Department,  Product Research and Development Department, Production Dept, Accounting Department and Personnel Department which are expressed as $SM, RD,ProD,AD,PerD$ respectively . The range of revenue of incident which forecasted by the company is divided into five levels which are \$10 million to \$20 million, \$20 million to \$30 million, \$30 million to \$40 million, \$40 million to \$50 million, and \$50 million to \$60 million. Therefore the frame of discernment of quantum is expressed $\Theta=\{Fir,Sec,Thi,Fou,Fif\}$. After the consideration of the order of these propositions, specific well-organized data of the process of judgment is shown in Table \ref{data2}

\begin{table}[h]\footnotesize
	\centering
	\caption{The calculation results parameter $d_{XP}(Q_{i},Q_{j})$ of  application4}
	\begin{spacing}{1.80}
		\begin{tabular}{c c c c c c }\hline
			\emph{Quantum evidence} &\multicolumn{4}{c}{Values of distance between QBPAs}\\\hline
			
			& $Fir$ & $Sec$ & $Thi$ & $Fou$ & $Fif$ \\ 
			$Evidence_{SM}$&  $0  $& $0.0477 $ &$ 0.0874$  & $0.0867 $ &$ 0.0773 $ \\ 
			$Evidence_{RD}$ & $0.0477  $&$ 0 $ & $0.0829 $ & $0.0973 $ & $0.0530  $\\ 
			$Evidence_{ProD}$&$ 0.0874 $ &$ 0.0829$  & $0  $& $0.0730 $ &$ 0.2892$  \\ 
			$Evidence_{AD}$&$0.0867 $ & $0.0973  $&$ 0.0730 $ & $0$ & $0.1055  $\\ 
			$Evidence_{PerD}$& $0.0773$  & $0.0530  $& $0.2892 $ & $0.1055$  & $0 $ \\ \hline
		\end{tabular}
	\end{spacing}
	%\label{tab:Margin_settings}
	\label{XP2}
\end{table}

What's more, the computation of $d_{XP}(Q_{i},Q_{j})$  and $d_{WB}(Q_{i},Q_{j})$ are enumerated in Table \ref{XP2} and \ref{WB2}. Then, $ Sim_{1}(Q_{i},Q_{j})$ and $ Sim_{2}(Q_{i},Q_{j})$ are obtained through utilizing formulas of similarity and two varieties of discrepancy frame of discernment of quantum  which are listed in Table \ref{Sim12} and \ref{Sim22}. Moreover, the weight that is taken into account order of each quantum evidence
\begin{figure}[h] %figure环境，h默认参数是可以浮动，不是固定在当前位置。如果要不浮动，你就可以使用大写float宏包的H参数，固定图片在当前位置，禁止浮动。
	\centering %使图片居中显示
	\includegraphics[width=0.7\textwidth]{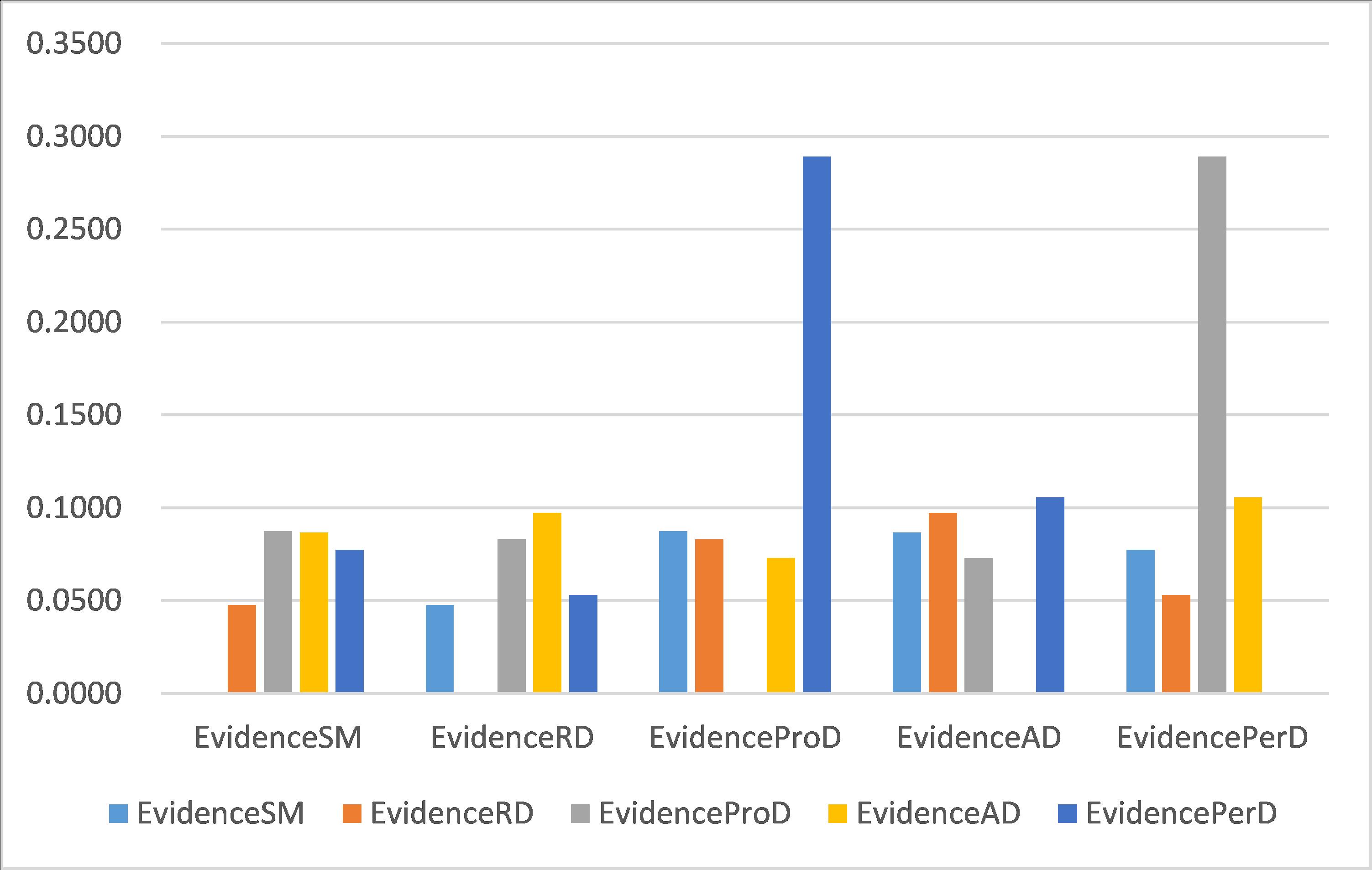} %中括号中的参数是设置图片充满文档的大小，你也可以使用小数来缩小图片的尺寸。
	\caption{The calculation result of parameter $Sim_{2}(Q_{i},Q_{j})$ of application1} %caption是用来给图片加上图题的
	\label{1} %这是添加标签，方便在文章中引用图片。
\end{figure}
is enumerated in Table \ref{wgt} which is calculated by a series of correctional combination rules raised by us of discernment of quantum. Finally, the comparison of results which are combined by making use of the proposed method and conventional approach in the quantum field and at the level of classic probability assignment are listed in Table \ref{quvalue} and \ref{value2} severally.

\begin{table}[h]\footnotesize
	\centering
	\caption{The calculation results parameter $d_{WB}(Q_{i},Q_{j})$ of  application4}
	\begin{spacing}{1.80}
		\begin{tabular}{c c c c c c }\hline
			Quantum evidence &\multicolumn{4}{c}{Values of distance between QBPAs}\\\hline
			
			& $Fir$ & $Sec$ & $Thi$ & $Fou$ & $Fif$ \\ 
			$Evidence_{SM}$&$ 0$ & $0.0604 $ &$ 0.1111$  & $0.1263 $ & $0.1047$  \\
			$Evidence_{RD}$ &$ 0.0604 $ &$ 0 $ & $0.1057 $ &$ 0.1358$  & $0.0357 $ \\ 
			$Evidence_{ProD}$& $0.1111 $ &$ 0.1057 $ &$ 0$  & $0.0841$  & $0.1174 $ \\ 
			$Evidence_{AD}$ &$ 0.1263$  & $0.1358$  & $0.0841$  & $0 $& $0.1188 $ \\ 
			$Evidence_{PerD}$&$ 0.1047 $ & $0.0357 $ & $0.1174 $ & $0.1188$  & $0  $\\ \hline
		\end{tabular}
	\end{spacing}
	%\label{tab:Margin_settings}
	\label{WB2}
\end{table}

By means of discussing the comparison of the traditional method of combination and the modified approach, arresting difference of probability assignment can be discovered. In the frame of quantum discernment, the discrepancy between disparate evidences are taken into account adequately which is contributed to relieve the misleading influence of extreme evidence to acquire more accurate and rational combined values. By observing the Table \ref{value2}, the obvious difference between proposed method and traditional rules of combination of
\begin{figure}[h] %figure环境，h默认参数是可以浮动，不是固定在当前位置。如果要不浮动，你就可以使用大写float宏包的H参数，固定图片在当前位置，禁止浮动。
	\centering %使图片居中显示
	\includegraphics[width=0.7\textwidth]{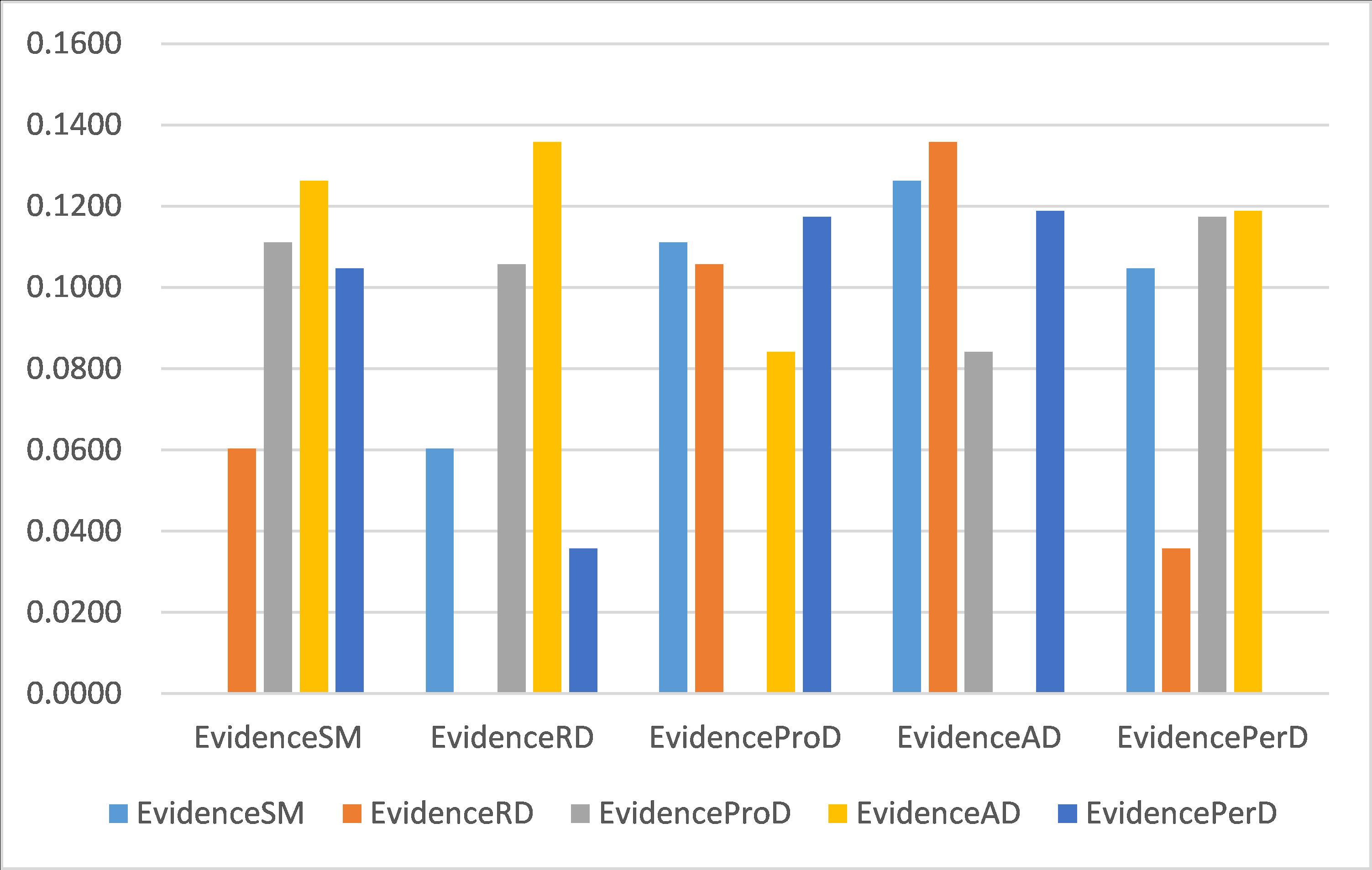} %中括号中的参数是设置图片充满文档的大小，你也可以使用小数来缩小图片的尺寸。
	\caption{The calculation result of parameter $Sim_{2}(Q_{i},Q_{j})$ of application1} %caption是用来给图片加上图题的
	\label{1} %这是添加标签，方便在文章中引用图片。
\end{figure}
final values can be told. By utilizing the traditional rule of combination, the second level of income-\$20 million to \$30 million is most likely to be achieved. However, in the ordinal frame of discernment of quantum, the probability assignment of accomplishing the first level of income-\$10 million to \$20 million is higher than others dramatically. Due to consideration and
\begin{table}[h]\footnotesize
	\centering
	\caption{The calculation results parameter $Sim_{1}(Q_{i},Q_{j})$ of application4}
	\begin{spacing}{1.80}
		\begin{tabular}{c c c c c c }\hline
			Quantum evidence &\multicolumn{4}{c}{Values of similarity  between QBPAs}\\\hline
			
			& $Fir$ & $Sec$ & $Thi$ & $Fou$ & $Fif$ \\ 
			$Evidence_{SM}$&$  1  $& $0.1030$  & $0.0862$  & $0.0943 $ & $0.0988 $ \\ 
			$Evidence_{RD}$ & $0.1030 $ &$ 1$  & $0.0942$  & $0.0899$  & $0.1248$  \\ 
			$Evidence_{ProD}$ & $0.0862$  &$ 0.0942$  &$ 1  $& $0.1191$  &$ 0.1014$  \\ 
			$Evidence_{AD}$ & $0.0943 $ & $0.0899 $ & $0.1191 $ &$ 1 $ &$ 0.0884 $ \\ 
			$Evidence_{PerD}$&$ 0.0988$  &$ 0.1248 $ &$ 0.1014$  &$ 0.0884$  & $1$ \\ \hline
		\end{tabular}
	\end{spacing}
	%\label{tab:Margin_settings}
	\label{Sim12}
\end{table}
discussion of the order of propositions of quantum evidence, two genres of important propositions $Fir$ and $Sec$ are neutralized on account of the weights of them are  similar and some relatively unimportant propositions are modified to further reduce their probability assignment which represents these events highly unlikely to happen. As a result, the proposed method which think over the order of evidence of quantum has better accuracy and unbiasedness in the direction of prediction.

\begin{figure}[h] %figure环境，h默认参数是可以浮动，不是固定在当前位置。如果要不浮动，你就可以使用大写float宏包的H参数，固定图片在当前位置，禁止浮动。
	\centering %使图片居中显示
	\includegraphics[width=0.7\textwidth]{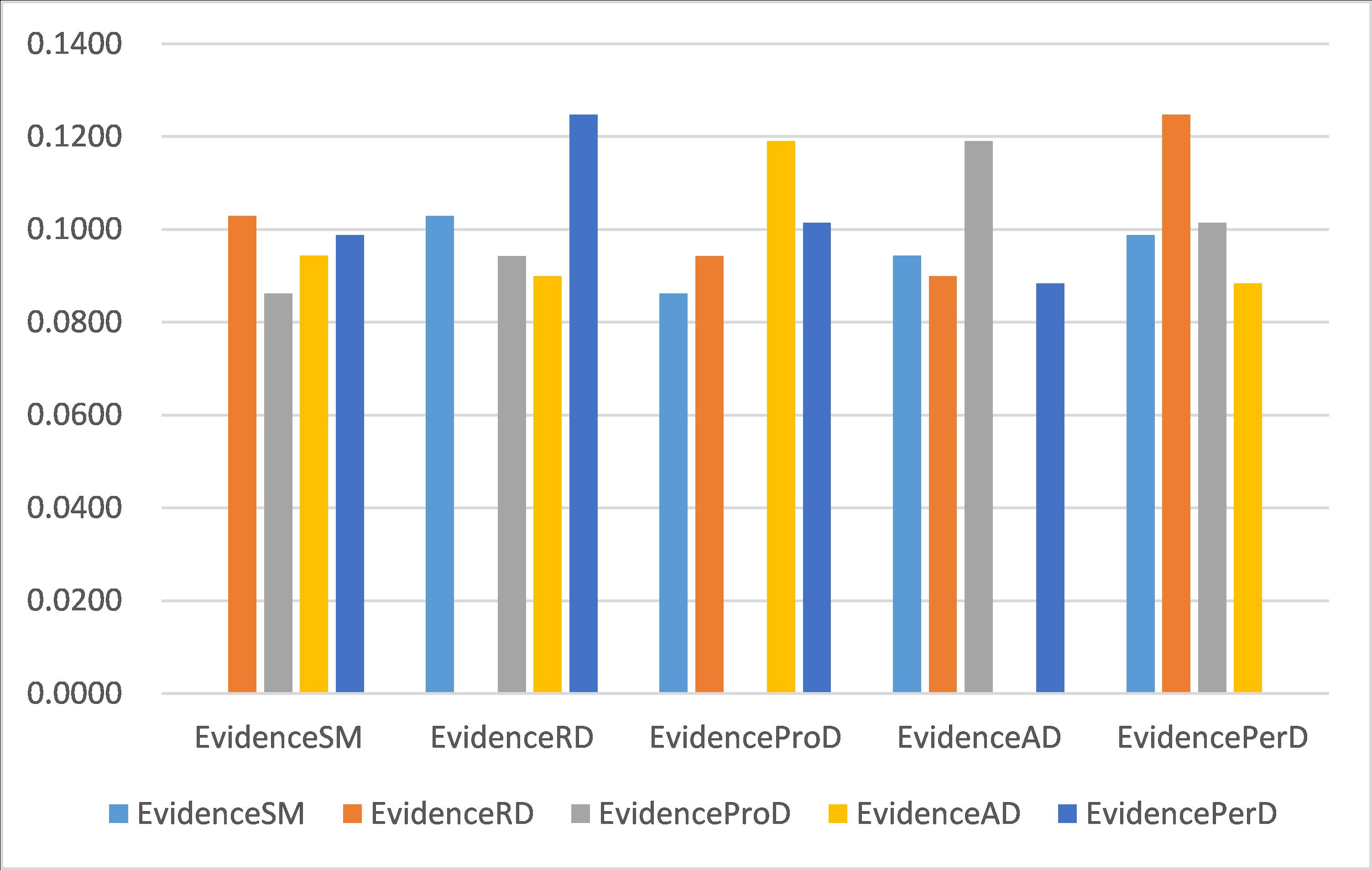} %中括号中的参数是设置图片充满文档的大小，你也可以使用小数来缩小图片的尺寸。
	\caption{The calculation result of parameter $Sim_{2}(Q_{i},Q_{j})$ of application1} %caption是用来给图片加上图题的
	\label{1} %这是添加标签，方便在文章中引用图片。
\end{figure}
\begin{table}[h]\footnotesize
	\centering
	\caption{The calculation results parameter $Sim_{2}(Q_{i},Q_{j})$ of application4}
	\begin{spacing}{1.80}
		\begin{tabular}{c c c c c c }\hline
			Quantum evidence &\multicolumn{4}{c}{Values of similarity  between QBPAs}\\\hline
			
			& $Fir$ & $Sec$ & $Thi$ & $Fou$ & $Fif$ \\ 
			$Evidence_{SM}$ &  $  1$  & $0.1104 $ & $0.1000$  & $0.0984$  & $0.1019$  \\ 
			$Evidence_{RD}$ & $0.1104 $ & $1 $& $0.1011 $ & $0.0962$  & $0.1126 $ \\ 
			$Evidence_{ProD}$ & $0.1000 $ &$ 0.1011$  & $1 $ & $0.1047$  & $0.0774$  \\ 
			$Evidence_{AD}$ & $0.0984 $ & $0.0962$  & $0.1047 $ &$ 1 $&$ 0.0972$  \\ 
			$Evidence_{PerD}$ & $0.1019 $ &$ 0.1126$  &$ 0.0774$  &$ 0.0972 $ & $1$  \\ \hline
		\end{tabular}
	\end{spacing}
	%\label{tab:Margin_settings}
	\label{Sim22}
\end{table}
\begin{figure}[h] %figure环境，h默认参数是可以浮动，不是固定在当前位置。如果要不浮动，你就可以使用大写float宏包的H参数，固定图片在当前位置，禁止浮动。
	\centering %使图片居中显示
	\includegraphics[width=0.7\textwidth]{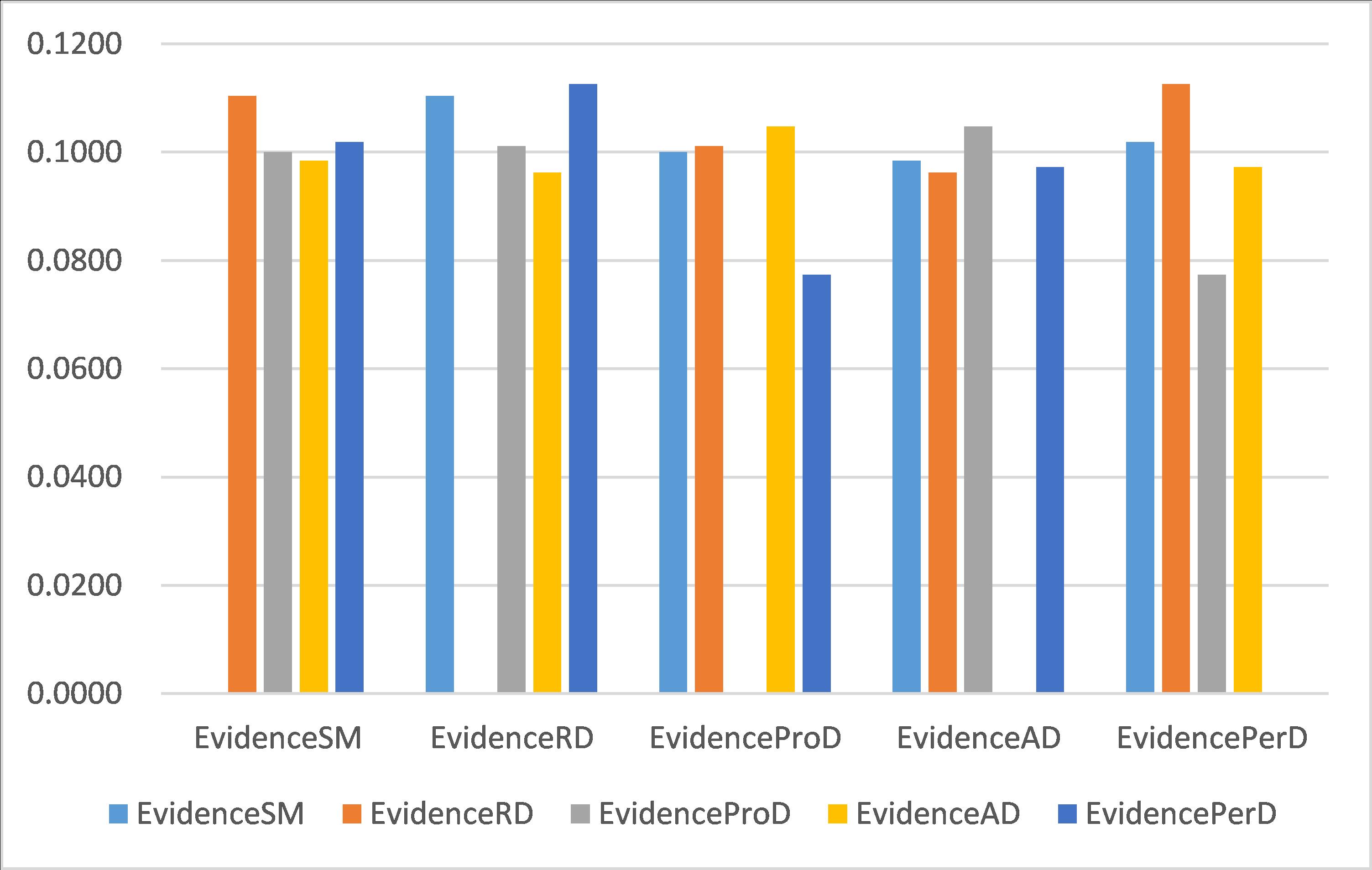} %中括号中的参数是设置图片充满文档的大小，你也可以使用小数来缩小图片的尺寸。
	\caption{The calculation result of parameter $Sim_{2}(Q_{i},Q_{j})$ of application1} %caption是用来给图片加上图题的
	\label{1} %这是添加标签，方便在文章中引用图片。
\end{figure}

\begin{table}[h]\footnotesize
	\centering
	\caption{The calculation results parameter $Wgt_{i}^{nor}$ of application 4}
	\begin{spacing}{1.80}
		\begin{tabular}{c c c c c c }\hline
			
			$Quantum \ \ evidence $	& $Fir$ & $Sec$ & $Thi$ & $Fou$ & $Fif$ \\  \hline
			$Wgt_{i}^{nor}$ & $0.1982 $& $0.2080$ & $0.1960$ & $0.1971$ &$ 0.2006 $\\ \hline
		\end{tabular}
	\end{spacing}
	%\label{tab:Margin_settings}
	\label{wgt}
\end{table}

\begin{table}[h]\footnotesize
	\centering
	\caption{The comparison of results combined between proposed method and traditional rule of combination in quantum field in application 4}
	\begin{spacing}{1.80}
		\begin{tabular}{c c c c c c}\hline
			
			$Proposition$ 	& $Fir$ & $Sec$ & $Thi$ & $Fou$ & $Fif$ \\  \hline
			$The \ \ improved \ \ combined \ \ values $ & $0.5443e^{2.5182i}$ & $0.8164e^{0.1408i}$& $0.0584e^{0.7536i}$ & $0.1155e^{1.7257i}$ & $0.1431e^{1.5691i}$\\ \hline
			$Proposition $	& $Fir$ & $Sec$ & $Thi$ & $Fou$ & $Fif$ \\  \hline
			$Combined \ \ values $& $0.7401e^{2.5269i}$& $0.6339e^{0.1205i}$& $0.1464e^{0.9514i}$& $0.1193e^{1.7262i}$  & $0.1216e^{1.5377i}$\\ \hline
		\end{tabular}
	\end{spacing}
	%\label{tab:Margin_settings}
	\label{quvalue}
	\label{}
	
\end{table}

\begin{table}[h]\footnotesize
	\centering
	\caption{The comparison of results combined between proposed method and traditional rule of combination in the form of classic probability assignment in application 4}
	\begin{spacing}{1.80}
		\begin{tabular}{c c c c c c}\hline
			
			$Proposition $	& $Fir$ & $Sec$ & $Thi$ & $Fou$ & $Fif$ \\  \hline
			$The \ \ improved \ \ combined \ \ values $& $0.2963 $& 0$.6665 $& $0.0034 $& $0.0133 $&$ 0.0205$   \\ \hline
			$Proposition $	& $Fir$ & $Sec$ & $Thi$ & $Fou$ & $Fif$ \\  \hline
			$Combined \ \ values $& $0.5477$ & $0.4018$ &$ 0.0214$ &$ 0.0142$ &$ 0.0148  $ \\ \hline
		\end{tabular}
	\end{spacing}
	%\label{tab:Margin_settings}
	\label{value2}
\end{table}
\begin{figure}[h] %figure环境，h默认参数是可以浮动，不是固定在当前位置。如果要不浮动，你就可以使用大写float宏包的H参数，固定图片在当前位置，禁止浮动。
	\centering %使图片居中显示
	\includegraphics[width=0.7\textwidth]{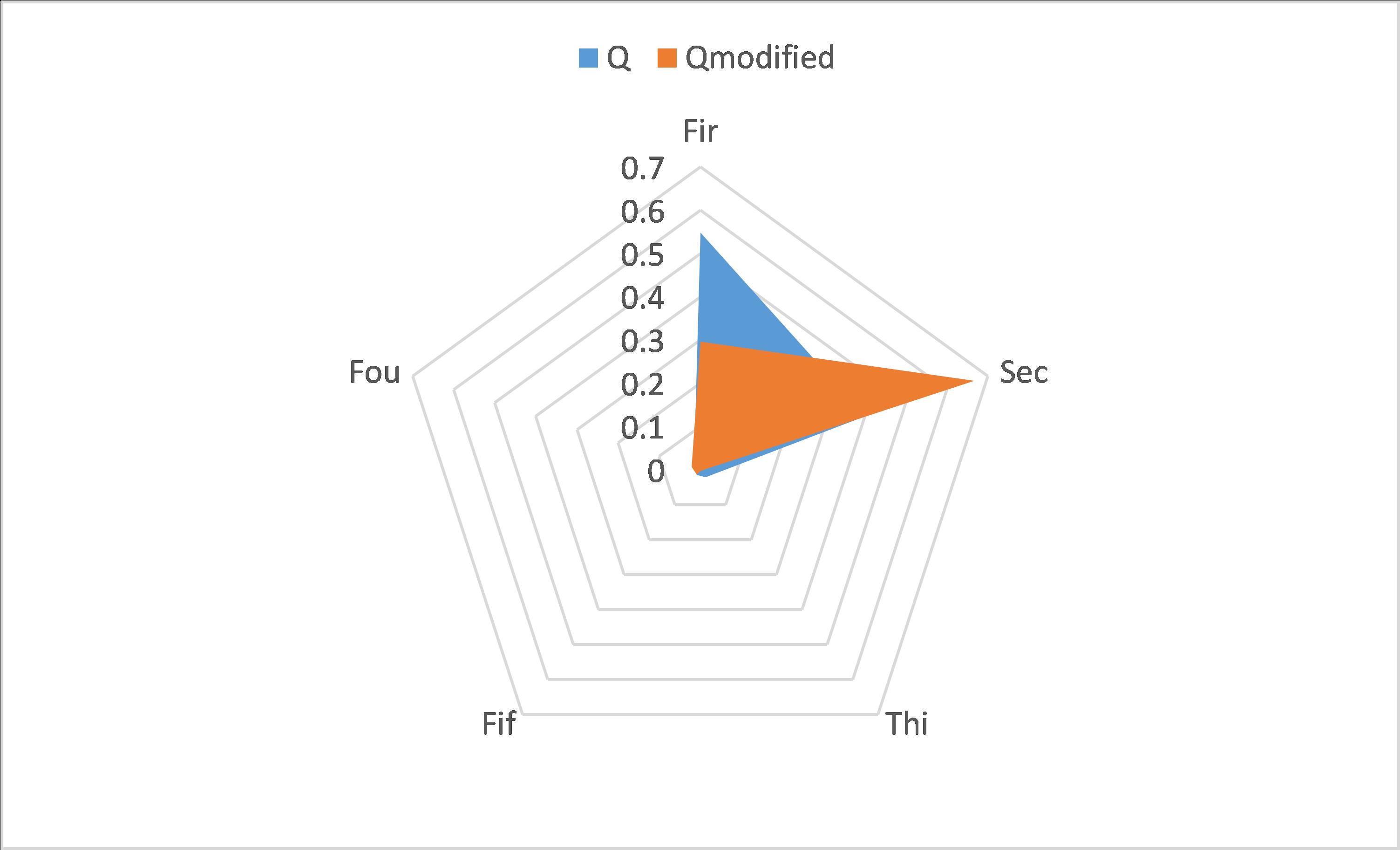} %中括号中的参数是设置图片充满文档的大小，你也可以使用小数来缩小图片的尺寸。
	\caption{The calculation result of parameter $Sim_{2}(Q_{i},Q_{j})$ of application 4} %caption是用来给图片加上图题的
	\label{1} %这是添加标签，方便在文章中引用图片。
\end{figure}
\clearpage
\section{Conclusion}
In this passage, a precursory method in combining ordinal quantum evidences is proposed, which provides an accurate and reasonable solution to alleviate uncertainty contained in quantum information. Two categories of difference measurement is designed to properly present underlying relationships among evidences. More than that, a customised degree of similarity is specially designed on the basis of figures displayed in the complex field. Those powerful tools offers a sufficient support in combining ordinal quantum evidences accordingly. The method proposed in this paper provides a completely view to dispose ordinal information given in the form of quantum and can be regarded as a superior, reasonable and completely new solution of combination of quantum evidence theory
\section*{Acknowledgment}
%The authors greatly appreciate the reviews' suggestions and the editor's encouragement.
 This research was funded by the Chongqing Overseas Scholars Innovation Program (No. cx2018077).

\bibliographystyle{elsarticle-num}
\bibliography{cite}
\end{document}